\documentclass[review]{elsarticle}

\usepackage{lineno,hyperref}
\modulolinenumbers[5]

\journal{Journal of \LaTeX\ Templates}

\usepackage{picins}
\usepackage{textcomp}
\usepackage{multirow}
\usepackage{hyperref}
\usepackage{microtype}
\usepackage{pgfplots, pgfplotstable, filecontents}
\pgfplotsset{compat=newest}
\usepackage{tikz}
\usetikzlibrary{patterns}
\usepackage{mathrsfs}

\usepackage{float}
\usepackage{graphicx}
\usepackage{natbib}
\usepackage{adjustbox}
\usepackage{subfigure}
\usepackage[linesnumbered,ruled]{algorithm2e}
\usepackage{setspace}
\SetKwComment{Comment}{$\///$\ }{}
\SetKwRepeat{Do}{do}{while}%
\newlength\mylen

\usepackage{nameref}
\usepackage[labelfont=bf,justification=raggedright,singlelinecheck=false]{caption}
\graphicspath{ {images/} }

\usepackage{multirow}
\usepackage{float}
\usepackage{tabularx}
\usepackage{booktabs}
\usepackage{ltablex}
\usepackage{threeparttable}
\usepackage{booktabs,makecell,siunitx}
\setcellgapes{5pt}
\usepackage{amssymb}
\usepackage{amsmath}

\begin{document}
\abovedisplayskip=0pt
\abovedisplayshortskip=0pt
\belowdisplayskip=0pt
\belowdisplayshortskip=0pt
\abovecaptionskip=0pt
\belowcaptionskip=0pt

\begin{frontmatter}

\title{Semi-supervised binary classification with latent distance learning}

\author[mymainaddress]{Imam Mustafa Kamal}
\ead{imamkamal@pusan.ac.kr}

\author[mysecondaryaddress]{Hyerim Bae\corref{mycorrespondingauthor}}
\cortext[mycorrespondingauthor]{Corresponding author}
\ead{hrbae@pusan.ac.kr}

\address[mymainaddress]{Institute of Intelligent Logistics Big Data, Pusan National University, Busan 46241, South Korea}
\address[mysecondaryaddress]{Major in Industrial Data Science \& Engineering, Department of Industrial Engineering, Pusan National University, Busan 46241, South Korea}

\begin{abstract}

Binary classification (BC) is a practical task that is ubiquitous in real-world problems, such as distinguishing healthy and unhealthy objects in biomedical diagnostics and defective and non-defective products in manufacturing inspections. Nonetheless, fully annotated data are commonly required to effectively solve this problem, and their collection by domain experts is a tedious and expensive procedure. In contrast to BC, several significant semi-supervised learning techniques that heavily rely on stochastic data augmentation techniques have been devised for solving multi-class classification. In this study, we demonstrate that the stochastic data augmentation technique is less suitable for solving typical BC problems because it can omit crucial features that strictly distinguish between positive and negative samples. To address this issue, we propose a new learning representation to solve the BC problem using a few labels with a random $k$-pair cross-distance learning mechanism. First, by harnessing a few labeled samples, the encoder network learns the projection of positive and negative samples in angular spaces to maximize and minimize their inter-class and intra-class distances, respectively. Second, the classifier learns to discriminate between positive and negative samples using on-the-fly labels generated based on the angular space and labeled samples to solve BC tasks. Extensive experiments were conducted using four real-world publicly available BC datasets. With few labels and without any data augmentation techniques, the proposed method outperformed state-of-the-art semi-supervised and self-supervised learning methods. Moreover, with 10\% labeling, our semi-supervised classifier could obtain competitive accuracy compared with a fully supervised setting.

\end{abstract}

\begin{keyword}
\texttt Binary classification \sep semi-supervised learning \sep angular distance \sep representation learning.
\MSC[2010] 00-01\sep  99-00
\end{keyword}

\end{frontmatter}


\section{Introduction}
\label{sec:introduction}

Binary classification (BC) is the task of classifying data samples into two classes, commonly called positive and negative sets. This is an essential task in machine learning problems, which are ubiquitous in real-world situations. In medical testing, data samples from observed patients are often categorized as healthy and unhealthy; an example is the detection between samples with an indication of cancer or normal. In quality control, samples of products generated during the manufacturing process can be classified as normal or defective. In information retrieval, samples of information on the Internet can be differentiated as normal and hoax (or containing spam, fraud, or malware). So far, several significant binary classifiers have been proposed, such as logistic regression, support vector machines, decision trees, random forests, naive Bayes, and (deep) neural networks. Nevertheless, their performances strictly rely on the existence of labels because they train the data to distinguish the positive and negative sets in a supervised manner. Furthermore, the amount of annotated (unlabeled) data nowadays (which is a digital and automated era) tends to be enormous, and the manual labeling of these data with the help of domain experts is a tedious and expensive task. Hence, one of the straightforward solutions, which does not neglect any (information) data samples, employs all of them with only a few numbered labels. The learning mechanism in this setting is called semi-supervised learning

Semi-supervised learning is a composite of supervised and unsupervised learning. In addition to unlabeled data, the algorithm provides certain supervision information \cite{vanEngelen2019ASO}. The proportion of labeled examples is usually relatively small compared to unlabeled examples. To the best of our knowledge, several semi-supervised methods have been introduced to solve multi-class classification problems \cite{NIPS2015_378a063b,DBLP:journals/corr/LaineA16,xie2020unsupervised}, and none of them explicitly devised a typical procedure for solving a BC problem. In multi-class classification, one of the prominent techniques for solving semi-supervised learning problems is the use of pseudo-labels. In the first stage, the classifier is trained using labeled sets. Next, the classifier predicts the labels of the unlabeled sets. However, only samples with predicted labels have a high probability (confidence level) that can be used in the second stage of the training process. For generalization performance, stochastic data augmentation is strictly required during training to avoid overfitting.

\begin{figure}[h]
	\centering
	\subfigure[malaria cells]{\includegraphics[width=0.25\linewidth]{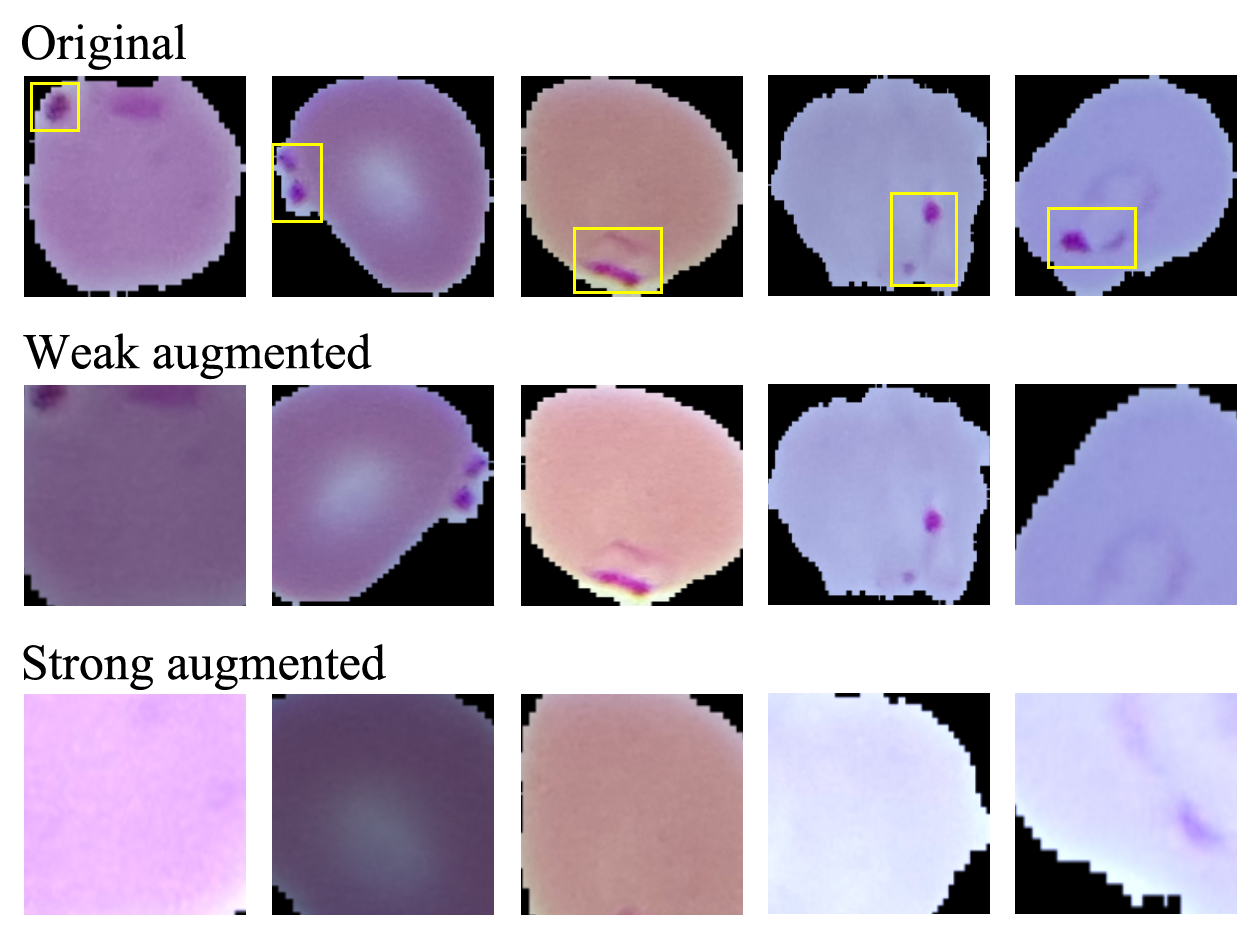}
		\label{cell_img}}
	\subfigure[brain tumors]{\includegraphics[width=0.25\linewidth]{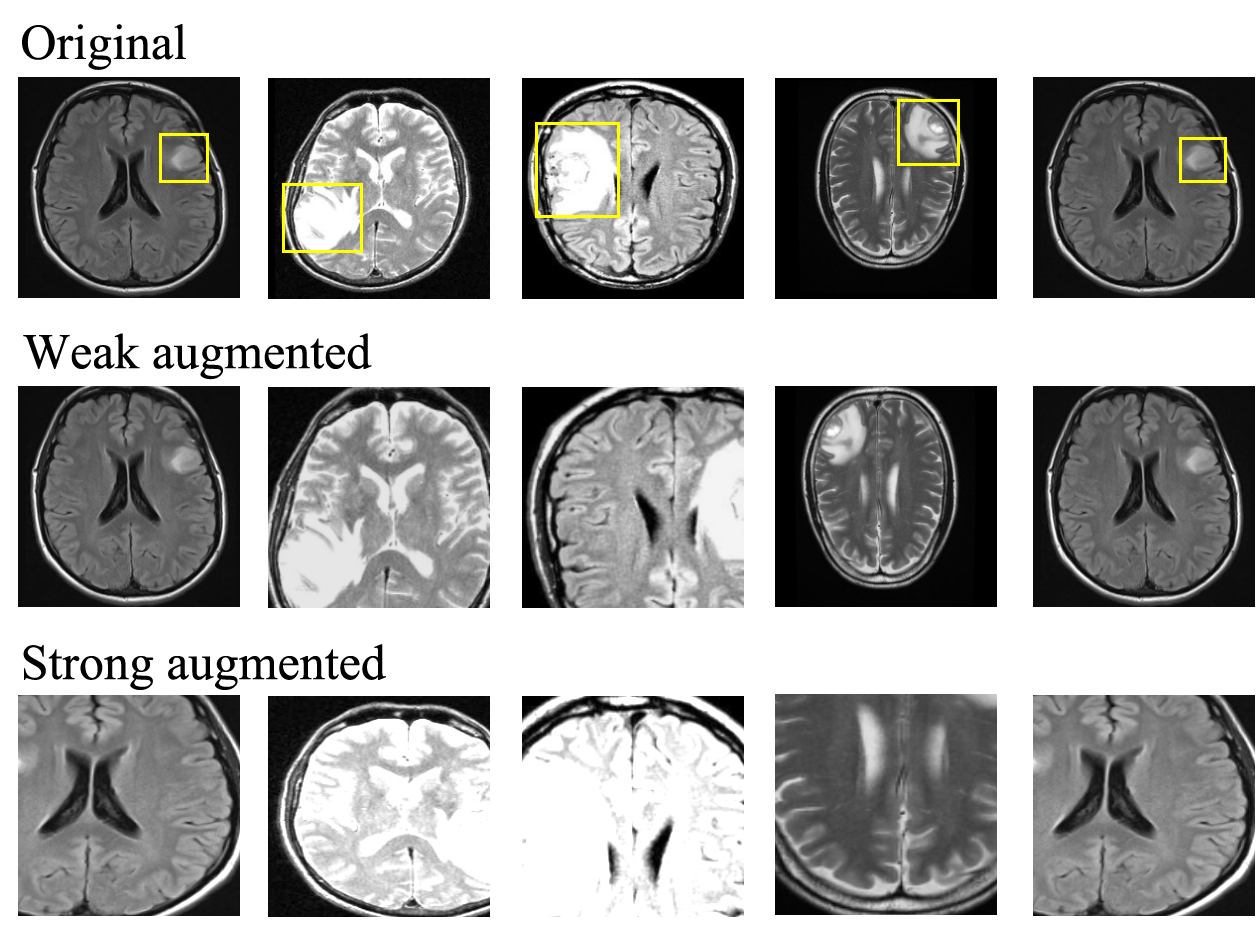}
		\label{tumor_img}}
	\subfigure[solar cells]{\includegraphics[width=0.25\linewidth]{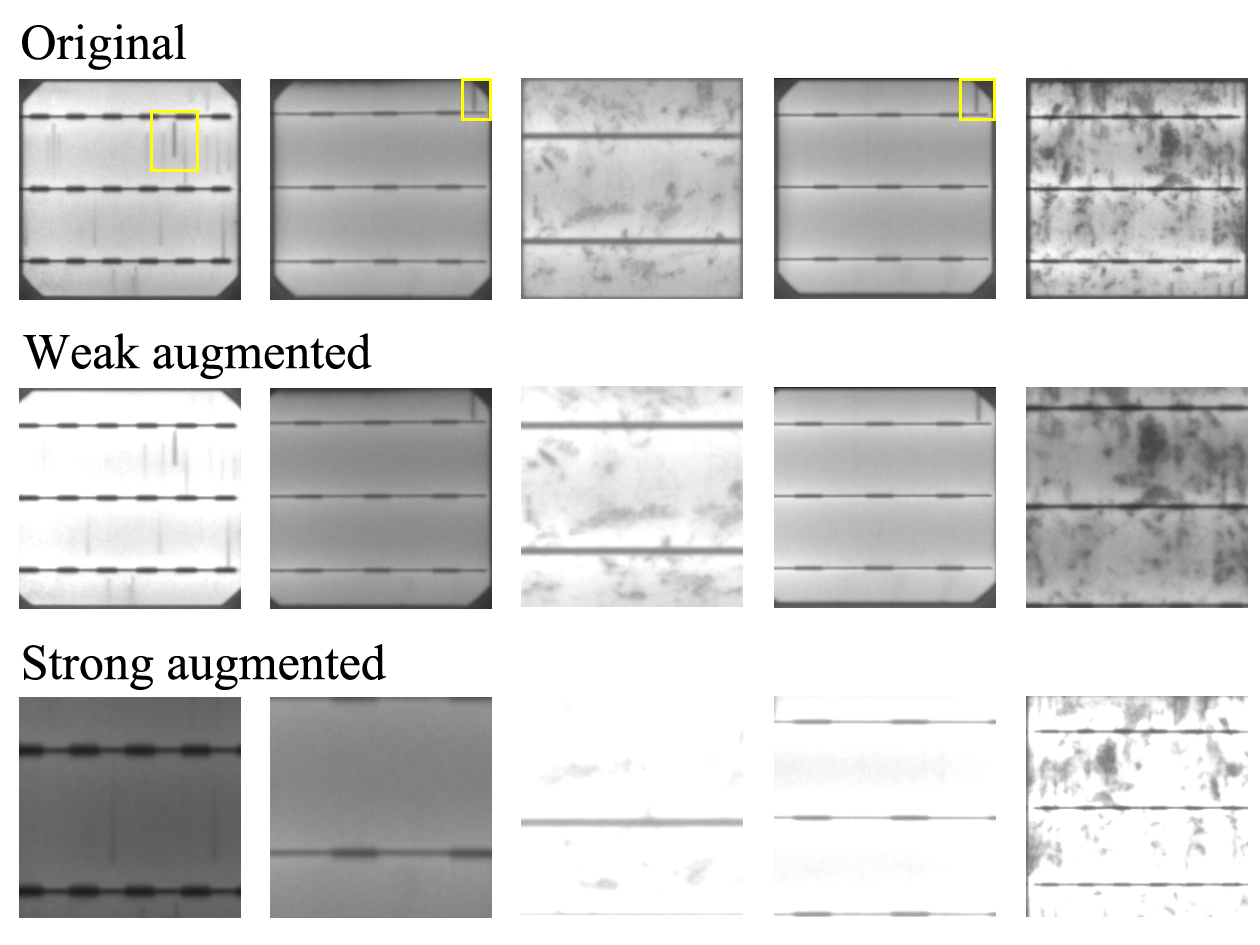}
		\label{solar_img}}
	\subfigure[crack surface]{\includegraphics[width=0.25\linewidth]{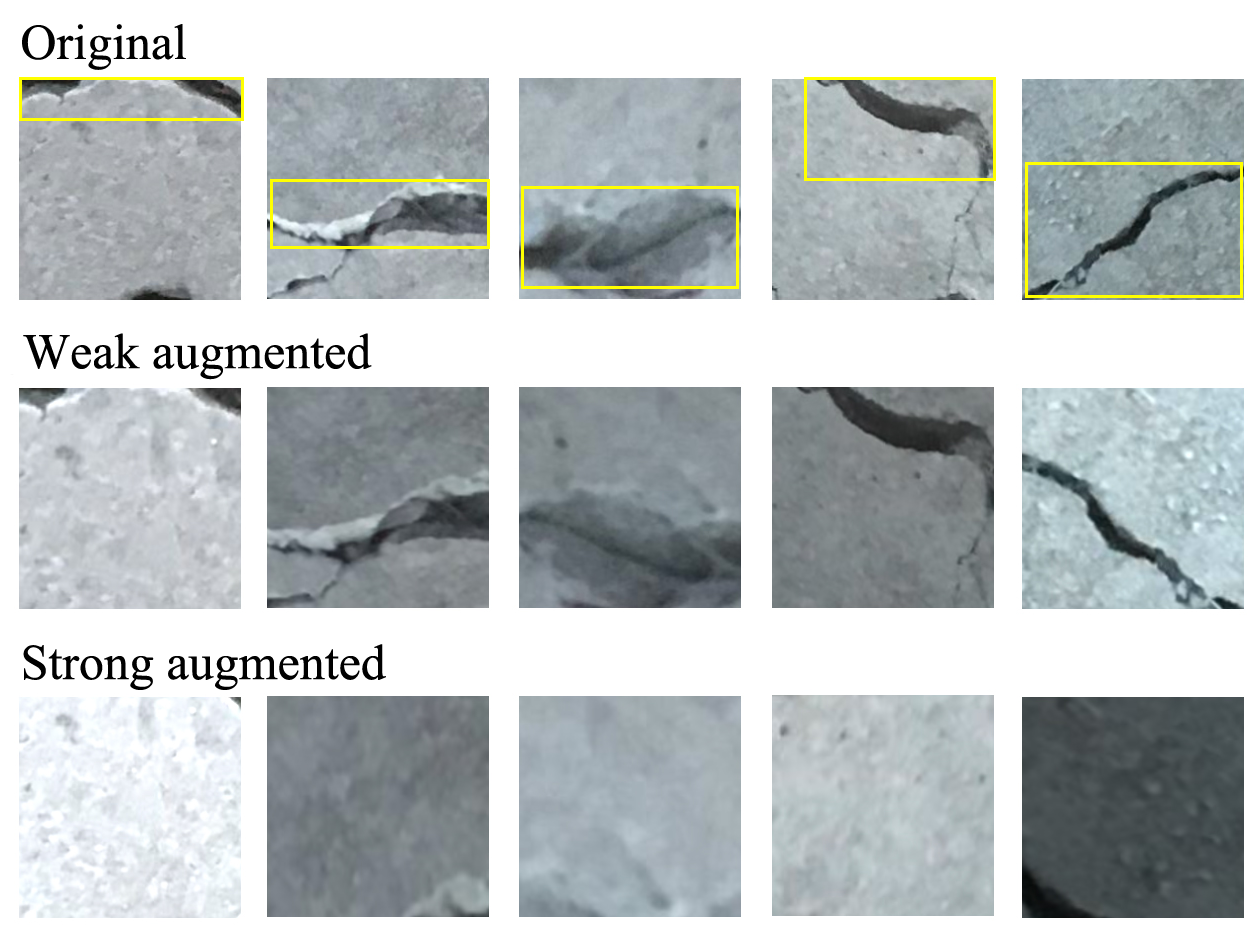}
		\label{crack_img}}
	\caption{Samples of BC datasets with their stochastic augmentation transformation. Original samples (first row), weak augmentation (second row), and strong augmentation (last row). The yellow box represents the important features reflected as positive samples; otherwise, they are negative.}
	\label{fig:data_examples}
\end{figure}

With the emergence of deep learning, self-supervised learning has been introduced as a new learning representation that does not rely on labels. Each sample is transformed into two different versions using strong and weak data augmentation processes, as shown in Fig. \ref{fig:data_examples}. Weak augmentation utilizes less aggressive transformation, such as applying random cropping, rotation, scaling, flipping, noise, and contrast to the image, whereas strong augmentation uses more aggressive transformation. The encoder model, which projects from high-dimensional to low-dimensional representation, is devised to distinguish them during training using contrastive loss \cite{NEURIPS2020_d89a66c7}. Nevertheless, some scholars demonstrated that with the help of a small portion of the labels (in a semi-supervised learning manner), the encoder can obtain a better separability performance compared with the self-supervision version \cite{assran2020supervision, dwibedi2021little}. Notably, the outcome of this representation learning is only an encoder network that is commonly used for downstream tasks, such as multi-class classification and BC. Using this technique, several semi-supervised classification tasks, particularly in multi-class classification (such as CIFAR, STL-10, and ImageNet), have been solved with remarkable and nearly similar results compared to the fully supervised setting, as demonstrated in the studies presented in \cite{dwibedi2021little,assran2021semi,assran2020supervision}. However, these approaches rely on stochastic data augmentation techniques. Strong data augmentation can reduce or even omit important features, as shown in Fig. \ref{fig:data_examples}. In BC datasets, the positive and negative samples are commonly distinguished by an important feature, which can be relatively subtle and trivial in some cases, as shown in Fig. \ref{fig:data_examples} \subref{cell_img} and \subref{solar_img}; thus, the feature can be removed while the images are augmented. Hence, transforming or removing this information can result in a misleading classification outcome because it can be detected in all negative samples

In this paper, we propose a new representation learning method that typically solves BC problems with only a few labels and without any data augmentation technique. First, with a few labeled samples, we devised metric learning to correctly project the distance between positive and negative samples in angular spaces. Thus, the distance of inter-class (positive-to-negative or negative-to-positive) and intra-class (positive-to-positive or negative-to-negative) is maximized and minimized, respectively, based on the angular distance. Second, we introduce a random $k$-paired cross-distance technique to generate on-the-fly labels for unlabeled data based on the supervision of the angular space. The binary classifier was trained using both labeled and unlabeled samples in a semi-supervised manner to obtain an accurate result. The proposed method was assessed using four publicly available datasets, demonstrating that it outperforms existing semi-supervised classifiers. Moreover, our method can obtain competitive accuracy compared to a fully supervised classifier with a small portion of the labels (10\%) and without any data augmentation technique

The remainder of this paper is organized as follows. Section \ref{sec:related_work} presents a recent literature review on semi-supervised classifiers. Section \ref{sec:method} provides a comprehensive description of the proposed method. Section \ref{sec:experiment} presents the experimental setting, results, and discussion. Finally, we summarize the conclusions, limitations, and future directions of this research in Section \ref{sec:conclusion}.

\section{Related works}
\label{sec:related_work}

Semi-supervised learning for classification with few labels is a popular area of research. However, most existing works focus on solving multi-class classification, where the number of classes is greater than two. These studies are still applicable to solving BC problems. Based on the data transformation used during training, we categorize the representation learning techniques used for semi-supervised classification into two main groups: random data perturbation and augmentation techniques. The data perturbation technique is mainly used for consistency regularization. Random perturbation is employed on unlabeled samples so that the model has a consistent prediction on a given unlabeled sample and its perturbed version. Rasmus et al. \cite{NIPS2015_378a063b} devised ladder networks with two encoders for projecting clean and perturbed samples to latent variables and a decoder to reconstruct both latent variables into the original (clean) version. However, the ladder network suffers from high computational cost, i.e., the number of computations required for one training iteration is approximately tripled. The $\pi$-model by Laine et al. \cite{DBLP:journals/corr/LaineA16} is a simplification of ladder networks, where the perturbed encoder is removed, and the same network is used to obtain the prediction for both clean and perturbed inputs. Miyato et al. \cite{miyato2018virtual} proposed the virtual adversarial training (VAT), i.e., a regularization technique that enhances the robustness of the model around each input data point against random and local perturbations. Nevertheless, applying the data perturbation transformation tends to be straightforward and causes the model to suffer from generalization errors. Xie et al. \cite{xie2020unsupervised} employed advanced data augmentation methods, such as AutoAugment, RandAugment, and Back Translation, as perturbations for consistency training. Similar to supervised learning, advanced data augmentation methods can provide additional advantages over simple augmentations and random perturbations for consistency regularization. The benefits of advanced augmentation are described as follows. First, it generates realistic augmented samples, making it safe to encourage consistency between predictions on the original and augmented samples. Second, it can generate a diverse set of examples to improve sample efficiency. Third, it can provide missing inductive biases for different tasks.

Inspired by the advantages of data augmentation, self-supervised representation learning aims to obtain robust representations of samples from raw data without expensive labels or annotations. This learning technique commonly uses two types of stochastic data augmentation as inputs, i.e., weak and strong. Accordingly, the model can distinguish the position of each sample in a manifold representation using contrastive loss. One of the earliest contrastive learning losses proposed was the information of noise-contrastive estimation (InfoNCE) loss \cite{DBLP:journals/corr/abs-1807-03748}. The similarity with contrastive learning (SimCLR) \cite{pmlr-v119-chen20j} extends the InfoNCE to learn representations that are invariant to image augmentation in a self-supervised manner by adding a projection head network. However, some scholars demonstrated that InfoNCE and SimCLR still provide unsatisfactory results because there is no supervision in projecting the latent variable in a manifold representation. Consequently, the semi-supervised nearest cross entropy (SuNCEt) \cite{assran2020supervision} employed a few labels to make an accurate manifold representation based on NCE and neighborhood component analysis. By using semi-supervision,  the nearest neighbor contrastive learning (NNCLR) \cite{dwibedi2021little}, which is based on the nearest neighbor technique, also demonstrated increased performance compared to existing contrastive learning methods, such as SimCLR and InfoNCE. It is worth noting that the aforementioned methods focused on defining pre-training tasks, which involved a surrogate task in a domain with ample weak supervision labels. The encoders trained to solve such tasks are expected to learn general features that might be useful for other downstream tasks requiring expensive annotations, such as multi-class classification or BC.

Another prominent approach for solving semi-supervised classification problems is the pseudo-labeling technique. The model is first trained on labeled data; then, at each training iteration, a portion of the unlabeled data is annotated (called a pseudo-label) using the trained model and added to the training set for the next iteration. Lee \cite{lee2013pseudo} proposed a network trained in a supervised manner with labeled and unlabeled data simultaneously. The pseudo-label is used as a label for unlabeled data based on a predefined threshold of confidence or probability. Li et al. \cite{Li2019NaiveSD} trained a classifier and used its outputs on unlabeled data as pseudo-labels. Then, they pre-trained the deep learning model with the pseudo-labeled data and fine-tuned it with the labeled data. The repetition of pseudo-labeling, pre-training, and fine-tuning is called naive semi-supervised deep learning. Berthelot et al. \cite{10555534542873454741} introduced MixMatch, which speculates low-entropy labels for data augmented unlabeled examples and mixes labeled and unlabeled data using MixUp. Berthelot et al. \cite{Berthelot2020ReMixMatch} also improved the MixMatch semi-supervised algorithm along the two directions of distribution alignment and augmentation anchoring, which together make the approach more data efficient than in prior works. Sohn et al. \cite{sohn2020fixmatch} devised FixMatch, which first generates pseudo-labels using the model predictions on weakly augmented unlabeled images. For a given image, the pseudo-label is retained only if the model produces a high confidence prediction. The model is then trained to predict the pseudo-label when fed with a strongly augmented version of the same image. Assran et al. \cite{assran2021semi} introduced a semi-supervised learning method by predicting view assignments with support samples (PAWS). This method trains a model to minimize consistency loss, which ensures that different views of the same unlabeled instance are assigned similar pseudo-labels.

Most of the aforementioned studies rely heavily on random perturbation and data augmentation, which can be problematic in typical BC datasets, as stated in Section \ref{sec:introduction}. Unlike previous studies, we propose a new research direction of semi-supervised classification based on a distance approach to further enhance the distance between positive and negative samples in manifold spaces. Moreover, inspired by the pseudo-labeling technique, we introduce a random $k$-pair cross-distance function to define an on-the-fly label for unlabeled samples during training.

\section{Methodology}
\label{sec:method}
The proposed method comprises two approaches, i.e., binary angular learning (BAL) and a semi-supervised binary classifier (SBC). The BAL learns the distance projection between the positive and negative samples in the angular spaces, whereas the SBC learns both labeled and unlabeled data to solve the BC problem. Below, we present a comprehensive problem formulation, detailed framework, and learning mechanism of the proposed framework.

\subsection{Preliminary}
\begin{figure}[H]
	\centering
	\includegraphics[width=0.67\linewidth]{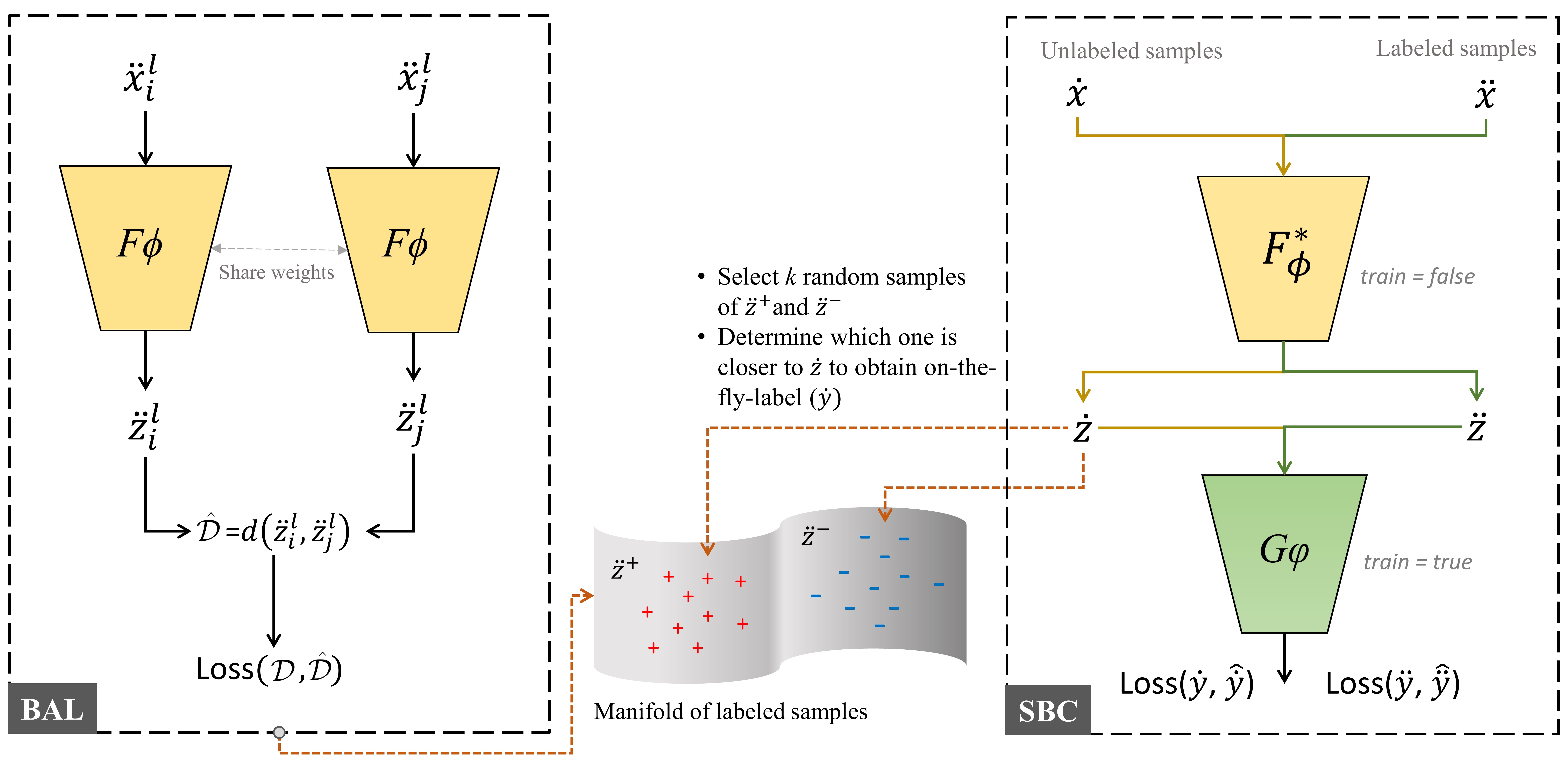}
	\caption{The visualization of the proposed framework. The left (BAL) represents metric learning to project the distance between the positive and negative samples in the angular spaces, which yields a binary manifold representation. The right (SBC) corresponds to the semi-supervised binary classifier to discriminate positive and negative classes based on the ground truth ($\ddot{y}$) and on-the-fly label ($\dot{y}$).}
	\label{fig:framework}
\end{figure}

We denote $X$ as a set of high-dimensional data consisting of labeled ($\ddot{X}$) and unlabeled ($\dot{X}$) data, and with dimensionality $p (X \in \mathbb{R}^p)$. Thus, $\ddot{X} = \{\ddot{x}^l_1,\ddot{x}^l_2,\ddot{x}^l_3,\dots,\ddot{x}^l_m\}$ and $\dot{X} = \{\dot{x}_{m+1},\dot{x}_{m+2},\dot{x}_{m+3},\dots,\dot{x}_n\}$, where $m$ and $n$ are the numbers of samples in $\ddot{X}$ and $X$, respectively. In semi-supervised classification, the number of labeled samples is typically less than or equal to the number of unlabeled samples $(m \leq (n-m))$. The label $l \in \{+,-\}$ consists of positive ($+$) and negative ($-$) classes as a standard BC problem. We denote $\neg l$ as a negation of label $l$; thus, if $l=+$, the $\neg l=-$, and vice versa. The encoder network $F_{\phi}$ projects $X$ into a low-dimensional representation $Z$, $Z \in \mathbb{R}^q$, where $q \ll p$. Hence, $\ddot{Z} = F_{\phi}(\ddot{X})$ or $\dot{Z} = F_{\phi}(\dot{X})$, where $\phi$ is a model parameter of $F$. Consequently, the binary classifier network $G_{\varphi}$ distinguishes whether $Z$ belongs to the positive $(+)$ or negative $(-)$ class. The overall proposed framework is illustrated in Fig. \ref{fig:framework}.

\subsection{Binary angular learning}
We aim to devise a binary manifold representation in which the positive and negative samples must be far away from each other. Accordingly, we employed the Siamese-like model \cite{Chicco2021} as metric learning to learn the projection between positive and negative samples. The Siamese network was trained in a weakly supervised manner, where the model did not directly access the label during training. However, each sample strictly requires the existence of a label to define the distance between classes in advance. Furthermore, in this study, the main characteristic of semi-supervised learning is that the amount of labeled data is relatively smaller than that of unlabeled data. Nevertheless, metric learning can still operate with small sample datasets, robustness to noisy data, high generalization to unseen categories, and capability of a dimensionality reduction model \cite{7407637}.

We assume that there are two labeled samples ${\ddot{x}}^{l}_i$ and ${\ddot{x}}^{l}_j$ from $\ddot{X}$ where $i \neq j$. The low-dimensional representation (vector), which is obtained from function $F_{\phi}$, is ${\ddot{z}}^{l}_i$ and ${\ddot{z}}^{l}_j$, respectively. Their cosine similarity ${s}({\ddot{z}}^{l}_i,{\ddot{z}}^{l}_j)$ can be expressed by Eq. \ref{eq:cos_similar}. The $s$ outcome is in the interval $[-1,1]$; thus, we could not employ it in our case, which is a binary problem. Therefore, we normalize Eq. \ref{eq:cos_similar}. The output in the interval $[0,1]$ after obtaining its angle $\theta$ and dividing it by $\pi={180}^{\circ}$ is called the angular distance ($d$), as denoted in Eq. \ref{eq:angular_dist}. If ${\ddot{z}}^{l}_i$ and ${\ddot{z}}^{l}_j$ are close to each other in the angular space or from the same class, ${d}({\ddot{z}}^{l}_i,{\ddot{z}}^{l}_j)=0$; otherwise, ${d}({\ddot{z}}^{l}_i,{\ddot{z}}^{l}_j)=1$. Therefore, in a supervised manner, with respect to the distance between the inputs ${\ddot{x}}_i$ and ${\ddot{x}}_j$, we can train $F_{\phi}$ to obtain the manifold distance representation of $\ddot{X}$, as illustrated in Fig. \ref{fig:framework} left.

\begin{equation}
	\label{eq:cos_similar}
	{s}({\ddot{z}}^{l}_i,{\ddot{z}}^{l}_j) = \frac{{\ddot{z}}^{l}_i \cdot {\ddot{z}}^{l}_j}{\ \| {\ddot{z}}^{l}_i \|\| {\ddot{z}}^{l}_j \|}
\end{equation}
\begin{equation}
	\label{eq:angular_dist}
	{d}({\ddot{z}}^{l}_i,{\ddot{z}}^{l}_j) = \frac{{\arccos}({s}({\ddot{z}}^{l}_i,{\ddot{z}}^{l}_j))}{\pi} = \frac{\theta_{{\ddot{z}}^{l}_i,{\ddot{z}}^{l}_j}}{\pi}
\end{equation}

\begin{center}
	\resizebox{0.6\linewidth}{!}{
		\centering
		\begin{algorithm}[H]
			\setstretch{0.75}
			\SetAlgoLined
			\caption{input mapping of binary angular distance}
			\label{algo:mapping_pair}
			\KwIn{\{${\ddot{x}}^{l}_1,{\ddot{x}}^{l}_2,{\ddot{x}}^{l}_3,\dots,{\ddot{x}}^{l}_{m}$\}}
			\KwOut{$\mathcal{X}$, $\mathcal{D}$} 
			$\mathcal{X}$ = [\phantom{0}] \DontPrintSemicolon \Comment*[r]{initialize $\mathcal{X}$ as an empty list}
			$\mathcal{D}$ = [\phantom{0}] \DontPrintSemicolon \Comment*[r]{initialize $\mathcal{D}$ as an empty list}
			\For{$i \gets 1$ \KwTo $m$}{
				\Do{${\ddot{x}}^{l}_i = {\ddot{x}}^{l}_j$}{
					${\ddot{x}}^{l}_j \gets$ \text{\textbf{random}}($j, l$) \DontPrintSemicolon \Comment*[r]{get random ${\ddot{x}}^{l}_j$}
					\DontPrintSemicolon \Comment*[r]{from the same class ($l$)}
				}
				$\mathcal{X}.$\text{\textbf{append}}($\{{\ddot{x}}^{l}_i,{\ddot{x}}^{l}_j\}$) \DontPrintSemicolon \Comment*[r]{a matching pair}
				$\mathcal{D}.$\text{\textbf{append}}(0) \DontPrintSemicolon \Comment*[r]{the angular distance = $0$}
				${\ddot{x}}^{l}_j \gets$ \text{\textbf{random}}($j,\neg l$) \DontPrintSemicolon \Comment*[r]{get random ${\ddot{x}}^{l}_j$}
				\DontPrintSemicolon \Comment*[r]{from different class ($\neg l$)}
				$\mathcal{X}.$\text{\textbf{append}}($\{{\ddot{x}}^{l}_i,{\ddot{x}}^{\neg l}_j\}$) \DontPrintSemicolon \Comment*[r]{a non-matching pair}
				$\mathcal{D}.$\text{\textbf{append}}(1) \DontPrintSemicolon \Comment*[r]{the angluar distance = $1$}
			}
		\end{algorithm}
	}
\end{center}

The learning process to obtain manifold distance representation is illustrated in Algorithms \ref{algo:mapping_pair} and \ref{algo:binary_distance}. In Algorithm \ref{algo:mapping_pair}, the labeled samples $\ddot{X}$ are first mapped to define their pairwise distance. The input of this algorithm is $\ddot{x}$ with its corresponding label $l$, whereas the output is a set of pairwise samples $\mathcal{X}$ and the angular distance ($\mathcal{D}$) of each pair in $\mathcal{X}$. As described in lines 7 and 10, the total number of data points is doubled ($2m$) because each sample is paired with the sample from both matching (same label) and non-matching (different label) classes. Next, $\mathcal{X}$ and $\mathcal{D}$ were used in the learning process, as shown in Algorithm \ref{algo:binary_distance}. For simplicity, we defined $\ddot{x}_j$ as a pair of $\ddot{x}_i$, which has either a matching ($l$) or non-matching ($\neg l$) label with $\ddot{x}_i$. The output of this algorithm is an optimal encoder model $F^*_{\phi}$, which is attained based on the validation sets; $\ddot{z}^l_i$ is obtained by model $F_{\phi}$ as a latent representation of $\ddot{x}^l_i$, similar to obtaining $\ddot{z}^l_j$ from $\ddot{x}^l_j$. Subsequently, the estimated distance ($\mathcal{\hat{D}}$) between them is calculated using Eq. \ref{eq:angular_dist}. Using the binary cross-entropy loss ($\mathcal{L}_{D}$) function (Eq. \ref{eq:bin_loss}), the discrepancy error between $\mathcal{D}$ and $\mathcal{\hat{D}}$ can be calculated to update the model parameter $\phi$. All the aforementioned processes are repeated within a predefined number of epochs ($NoEpoch$), as shown in lines 1 to 6. 

\begin{equation}
	\label{eq:bin_loss} 
	\mathcal{L}_{D} = -\frac{1}{2m}\Sigma_{i=1}^{2m} \mathcal{D} \cdot \log \mathcal{\hat{D}} + (1 - \mathcal{D}) \cdot \log (1 - \mathcal{\hat{D}})
\end{equation}

\begin{center}     
	\resizebox{0.6\linewidth}{!}{
		\centering
		\begin{algorithm}[H]
			\setstretch{0.75}
			\SetAlgoLined
			\caption{training procedure of BAL}
			\label{algo:binary_distance}
			\KwIn{$\mathcal{X}$ = \{${\{\ddot{x}^l_i,\ddot{x}^l_j\}}_1,{\{\ddot{x}^l_i,\ddot{x}^{\neg l}_j\}}_2,{\{\ddot{x}^l_i,\ddot{x}^{l}_j\}}_3,\dots,{\{\ddot{x}^l_i,\ddot{x}^{\neg l}_j\}}_{2m}\}$, $\mathcal{D}$} 
			\KwOut{$F^*_{{\phi}}$}
			\For{$i \gets 1$ \KwTo $NoEpoch$}{
				$\ddot{z}^l_i = F_{{\phi}_i}(\ddot{x}^l_i)$ \DontPrintSemicolon \Comment*[r]{encoder extracts $\ddot{z}^l_i$}
				$\ddot{z}^l_j = F_{{\phi}_i}(\ddot{x}^l_j)$ \DontPrintSemicolon \Comment*[r]{encoder extracts $\ddot{z}^l_j$}
				$\mathcal{\hat{D}} = d(\ddot{z}^l_i,\ddot{z}^l_j)$\DontPrintSemicolon \Comment*[r]{estimate the distance, Eq. \ref{eq:angular_dist}}
				$\mathcal{L}_{{\phi}_i} =$\text{\textbf{loss\_function}($\mathcal{D},\mathcal{\hat{D}}$)} \DontPrintSemicolon \Comment*[r]{calculate loss, Eq. \ref{eq:bin_loss}}
				${\phi}_i \gets {\phi}_i - \eta \nabla_{{\phi}_i}{\mathcal{L}}_{{\phi}_i}$ \DontPrintSemicolon \Comment*[r]{update encoder model parameter}
			}
		\end{algorithm}
	}
\end{center}

\subsection{Semi-supervised binary classifier}
In this section, the BC model is trained using both labeled and unlabeled samples. The training procedure is shown in Algorithm \ref{algo:semi_classifier}. We assume $\ddot{y}=\dot{y}=0$ as the positive class ($+$) and $\ddot{y}=\dot{y}=1$ as the negative class ($-$). Given the optimum encoder model $F^*_{\phi}$, the input $\ddot{x}$ is projected onto the low-dimensional vector representation $\ddot{z}$. A similar procedure can be applied to obtain $\dot{z}$ from the unlabeled input $\dot{x}$. We denote $\dot{y}$ by the on-the-fly label of $\dot{z}$ (or $\dot{x}$). Its value is obtained using closeness of normalized distance ($d_{n}$) between $\dot{z}$ and the $k-$random samples of $\ddot{z}^+$ and $\ddot{z}^-$, as expressed in Eqs. \ref{eq:prob_k_cross} and \ref{eq:k_cross}. It should be noted that in Eq. \ref{eq:prob_k_cross}, $\neg l$ represents the negation of label $l$; thus, if $l=+$, the $\neg l = -$, and vice versa. Hence, if the distance from $\dot{z}$ is closer to the $\ddot{z}^+$ than $\ddot{z}^-$, the $d_{n}(\dot{z},\ddot{z}^+)$ value will be smaller than $d_{n}(\dot{z},\ddot{z}^-)$, and thus $\dot{y}=0$. Conversely, if the distance from $\dot{z}$ is farther to the $\ddot{z}^+$ than $\ddot{z}^-$, the $d_{n}(\dot{z},\ddot{z}^+)$ value will be higher than $d_{n}(\dot{z},\ddot{z}^-)$, and thus $\dot{y}=1$. $d_n$ may be changed for each iteration because the $k$ samples of $\ddot{z}^+$ and $\dot{z}^-$ are selected randomly; hence, $\dot{y}$ is called an on-the-fly label. In Algorithm \ref{algo:semi_classifier}, we conduct two consecutive updates of the gradient for each epoch for the labeled ($\ddot{y}$) and unlabeled ($\dot{y}$) samples. Given that $\dot{y}$ is an on-the-fly label that is not guaranteed to be 100\% correct, the loss calculation between $\ddot{y}$ and $\hat{\ddot{y}}$ is executed first because we have an optimistic direction to reduce the gradient in updating parameter $\varphi$. The loss function in lines 6 and 9 is the binary cross-entropy, as expressed in Eqs. \ref{eq:bin_loss3} ($\mathcal{L}_{B1}$) and \ref{eq:bin_loss2} ($\mathcal{L}_{B2}$). In practice, the model is trained using a predefined number of batches, and the optimum classifier model $G_{\varphi}$ is obtained based on the validation set.

\begin{equation}
	\label{eq:prob_k_cross} 
	d_{n}(\dot{z},\ddot{z}^l) = \frac{d(\dot{z}, \ddot{z}^l)}{d(\dot{z}, \ddot{z}^l) + d(\dot{z}, \ddot{z}^{\neg l})}
\end{equation}

\begin{equation}
	\label{eq:k_cross}
	\dot{y}_i= 
	\begin{cases}
		0,& \text{if } \Sigma_{j=1}^{k} d_{n}(\dot{z}_i,\ddot{z}^+_j) \leq \Sigma_{j=1}^{k} d_{n}(\dot{z}_i,\ddot{z}^-_j)\\
		1,              & \text{otherwise}
	\end{cases}
\end{equation}

\begin{equation}
	\label{eq:bin_loss3} 
	\mathcal{L}_{B1} = -\frac{1}{m}\Sigma_{i=1}^{m} \ddot{y} \cdot \log \hat{\ddot{y}} + (1 - \ddot{y}) \cdot \log (1 - \hat{\ddot{y}})
\end{equation}
\begin{equation}
	\label{eq:bin_loss2} 
	\mathcal{L}_{B2} = -\frac{1}{(n-m)}\Sigma_{i=(m+1)}^{n} \dot{y} \cdot \log \hat{\dot{y}} + (1 - \dot{y}) \cdot \log (1 - \hat{\dot{y}})
\end{equation}

\begin{center}     
	\resizebox{0.6\linewidth}{!}{
		\centering
		\begin{algorithm}[H]
			\setstretch{0.75}
			\SetAlgoLined
			\caption{training procedure of SBC}
			\label{algo:semi_classifier}
			\KwIn{$\{\{\ddot{x},\ddot{y}\}_1,\{\ddot{x},\ddot{y}\}_2,\{\ddot{x},\ddot{y}\}_3,\dots,\{\ddot{x},\ddot{y}\}_m\}, \{\dot{x}_{m+1},\dot{x}_{m+2},\dot{x}_{m+3},\dots,\dot{x}_n\}$} 
			\For{$i \gets 1$ \KwTo $NoEpoch$}{
				$\ddot{z} = F^*_{\phi}(\ddot{x})$ \DontPrintSemicolon \Comment*[r]{pre-trained encoder extracts $\ddot{z}$}
				$\dot{z} = F^*_{\phi}(\dot{x})$ \DontPrintSemicolon \Comment*[r]{pre-trained encoder extracts $\dot{z}$}
				$\dot{y} =$ Eq. \ref{eq:k_cross} \DontPrintSemicolon \Comment*[r]{obtain on-the-fly label based on} \DontPrintSemicolon \Comment*[r]{the closeness of $\dot{z}$ to $k-$random $\ddot{z}^+$ and $\ddot{z}^-$}
				$\hat{\ddot{y}} = G_{{\varphi}_i}(\ddot{z})$ \DontPrintSemicolon \Comment*[r]{estimate the label}
				$\mathcal{L}_{{\varphi}_i} =$\text{\textbf{loss\_function}($\ddot{y},\hat{\ddot{y}}$)} \DontPrintSemicolon \Comment*[r]{calculate loss, Eq. \ref{eq:bin_loss3}}
				${\varphi}_i \gets {\varphi}_i - \eta \nabla_{{\varphi}_i}{\mathcal{L}}_{{\varphi}_i}$ \DontPrintSemicolon \Comment*[r]{update classifier parameter}
				\DontPrintSemicolon \Comment*[r]{optimistically}
				$\hat{\dot{y}} = G_{{\varphi}_i}(\dot{z})$ \DontPrintSemicolon \Comment*[r]{estimate the label}
				$\mathcal{L}_{{\varphi}_i} =$\text{\textbf{loss\_function}($\dot{y},\hat{\dot{y}}$)} \DontPrintSemicolon \Comment*[r]{calculate loss, Eq. \ref{eq:bin_loss2}}
				${\varphi}_i \gets {\varphi}_i - \eta \nabla_{{\varphi}_i}{\mathcal{L}}_{{\varphi}_i}$ \DontPrintSemicolon \Comment*[r]{update classifier parameter}
				\Comment*[r]{less optimistically}
			}
		\end{algorithm}
	}
\end{center}

\section{Result and discussion}
\label{sec:experiment}

\subsection{Data and setting}
\begin{figure}[h]
	\centering
	\subfigure[MC]{\includegraphics[width=0.3\linewidth]{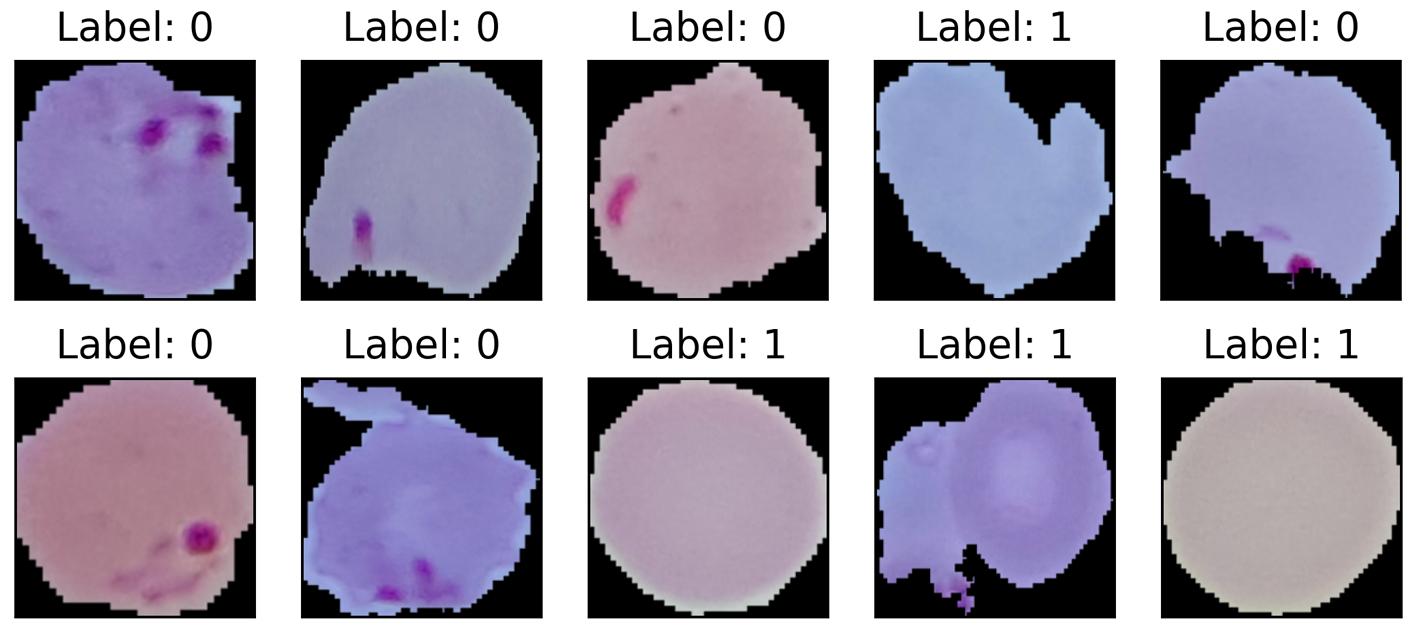}
		\label{cell_overview}}
	\subfigure[BT]{\includegraphics[width=0.3\linewidth]{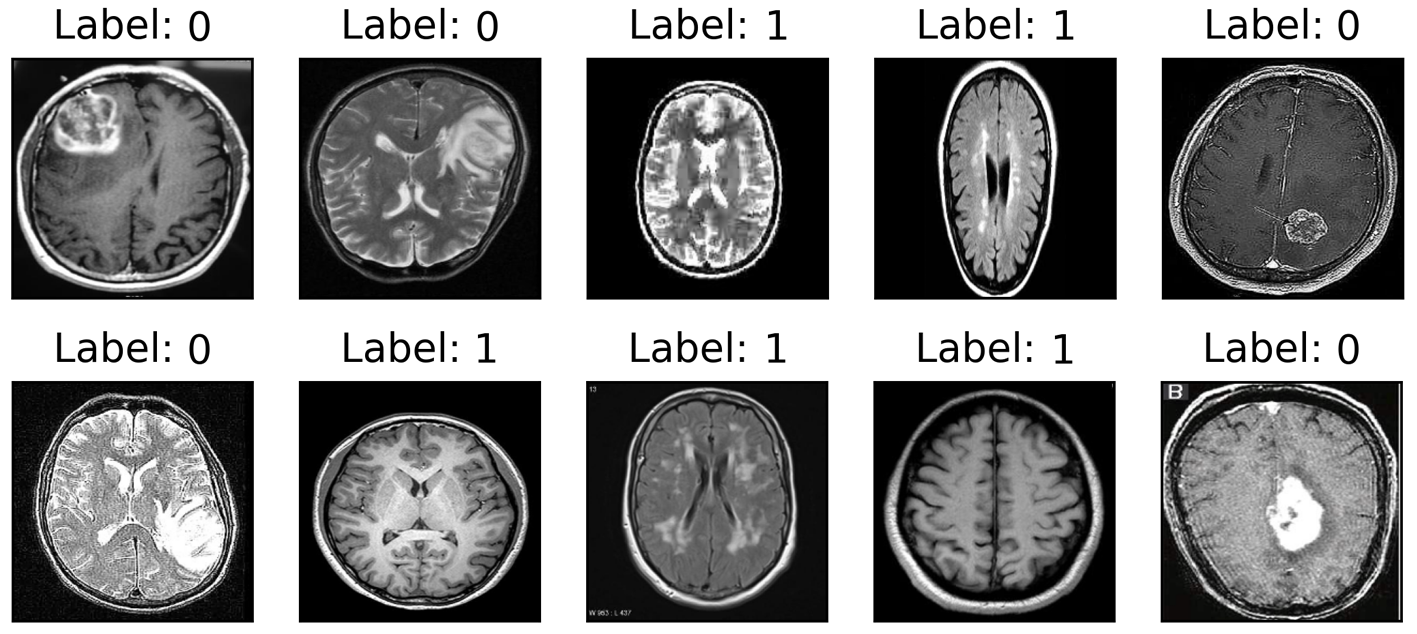}
		\label{tumor_overview}}
	\subfigure[SC]{\includegraphics[width=0.3\linewidth]{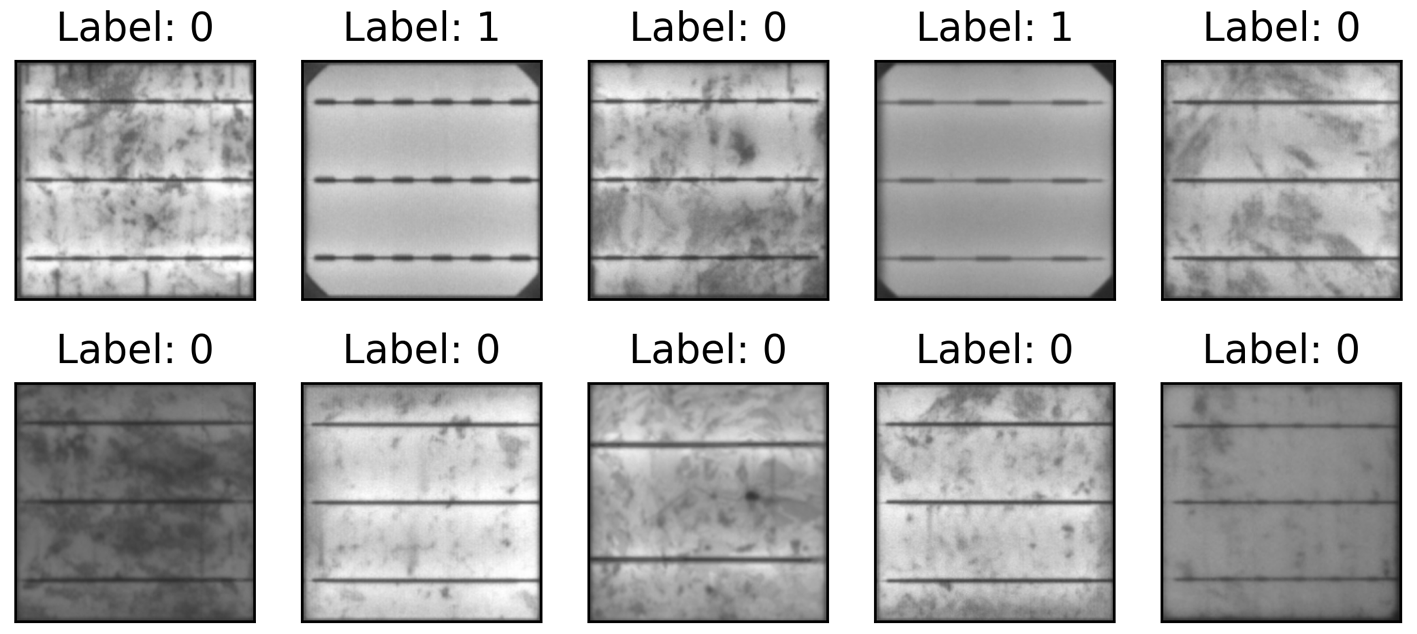}
		\label{solar_overview}}
	\subfigure[CS]{\includegraphics[width=0.3\linewidth]{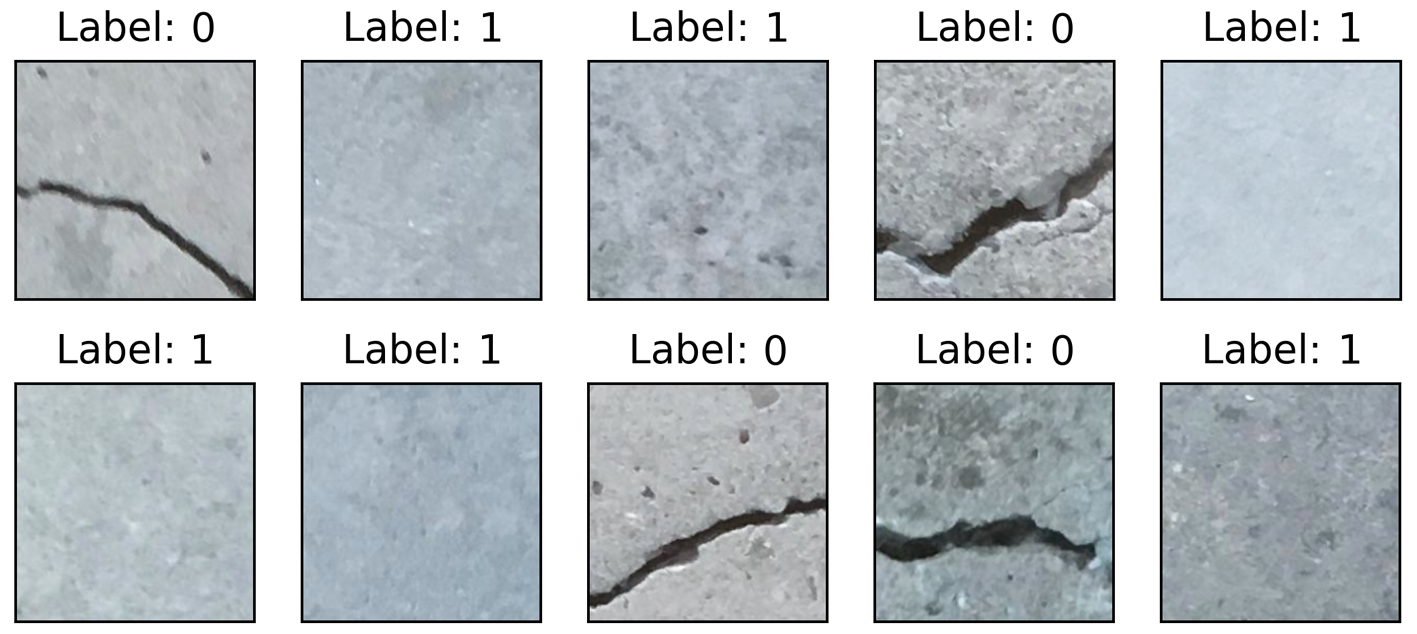}
		\label{crack_overview}}
	\caption{Samples of images from malaria cells (MC), brain tumors (BT), solar cells (ST), and crack surface (CS) datasets.}
	\label{fig:data_overview}
\end{figure}

We evaluated the proposed method using four publicly available datasets to solve real-world BC tasks: malaria cells (MC) \cite{rajaraman2018pre}, brain tumors (BT) \cite{ahmedhamada0}, solar cells (SC) \cite{DEITSCH2019455}, and crack surface (CS) \cite{s22103662} datasets (shown in Fig. \ref{fig:data_overview}). The MC dataset originated from the official National Institutes of Health (NIH) website containing $27,558$ cell images with equal instances of parasitized and uninfected cells from thin blood smear slide images of segmented cells. The BT dataset was obtained from a magnetic resonance imaging (MRI) technique consisting of $3,000$ samples with equal instances of a tumorous and healthy brain. The SC dataset contained $2,624$ sample images of functional ($1,550$) and defective ($1,074$) solar cells with varying degrees of degradation extracted from various solar modules. The CS contained $20,000$ color images with equal negative and positive crack samples collected from various Metu campus buildings. The image dimensionality of each dataset varies; thus, we resized them to $224 \times 224$ pixels and normalized them in the interval $[0,1]$. Each dataset was divided into 10-fold cross-validation to avoid bias. We split each dataset into an 80\%:20\% ratio for the training and testing sets, respectively; to obtain an optimal classifier, we employed a validation set that was subtracted by 20\% from the training set.
We conducted extensive trial-and-error experiments, including hyperparameter tuning, to obtain an optimum network architecture by strictly considering both accuracy and simplicity. Consequently, we obtained a VGG-like model as an encoder network ($F_{\phi}$) and a simple neural network with two layers as a binary classifier network ($G_{\varphi}$). The detailed architectures of these networks are presented in Table \ref{tab:net_architecture}. Moreover, we employed one of the prominent deep learning models, namely VGG-16, with ImageNet weights as our additional experiments to evaluate the model’s transfer learning capability. Hereafter, we denote SemBC as the proposed model name. To assess the effectiveness of the proposed method, we compared it with recent state-of-the-art semi-supervised classifiers such as SimCLR, NNCLR, SunCet, Entropy, FixMatch, and PAWS. The Entropy model represents a standard encoder model to learn manifold representation using only labeled data and binary cross-entropy as a loss function. Next, it was used as the backbone to train the binary classifier network ($G_{\varphi}$). The learning procedures of SimCLR, NNCLR, SunCet, and PAWS were described previously in Section \ref{sec:related_work}. To perform a fair comparison, we utilized $F_{\phi}$ and $G_{\varphi}$ with the same architecture and parameters in all models.

\begin{table}
	\centering
	\caption{Detailed network architecture of $F_{\phi}$ and $G_{\varphi}$.}
	\label{tab:net_architecture}
	\resizebox{1\linewidth}{!}{
		\begin{tabular}{l}		
			\toprule
			\textbf{Encoder ($F_{\phi}$)}:\\
			$Conv(64,3,Relu)-Conv(64,3,Relu)-MaxPool(3)-Conv(128,3,Relu,L1-L2)-Conv(128,3,Relu,L1-L2)-$ \\
			$MaxPool(3)-BatchNorm-Conv(256,3,Relu,L1-L2)-Conv(256,3,Relu,L1-L2)-MaxPool(3)-BatchNorm-$\\
			$Conv(512,3,Relu,L1-L2)-Conv(512,3,Relu,L1-L2)-MaxPool(2)-BatchNorm-GlobAvgPool-Dense(512,Relu,L1-L2)-$\\
			$Dense(512,Relu,L1-L2)-Dense(256)-L2Norm.$\\
			\midrule
			\textbf{Binary classifier ($G_{\varphi}$)}:\\
			$Dense(80,Relu,L1-L2)-BatchNorm-Dense(20,Relu,L1-L2)-BatchNorm-Dense(1,Sigmoid).$\\
			\bottomrule
		\end{tabular}
	}
\end{table}
In Table \ref{tab:net_architecture}, $Conv(a,b,c)$ represents a convolutional layer with  $a$ number of filters, kernel size $b$, and activation function $c$. $MaxPool(a)$ is a two-dimensional pooling layer of size $a \times a$. $BatchNorm$ is a batch normalization layer, and $GlobAvgPool$ is a two-dimensional global average pooling layer. $Dense(a,b,c)$ corresponds to a dense layer with unit neuron $a$, activation function $b$, and $c$ as $L1$ and $L2$ regularization. $L2Norm$ is a $L2$ normalization layer. We utilized stochastic gradient descent (SGD) with a learning rate of 0.01, momentum of 0.9, and clip normalization of 1.0 to optimize $F_{\phi}$. The Adam optimization method was used in the $G_{\varphi}$ with a learning rate of 0.001. We trained both $F_{\phi}$ and $G_{\varphi}$ for 30 epochs with a batch size of 32 in the MC, SC, and CS datasets (batch size of BT = 16). Our experiment was conducted using a TITAN RTX GPU, Python ver. 3.7.6, Tensorflow ver. 2.8.0, and Keras ver. 2.8.0.

\subsection{Classification performance}

\begin{table}[h]
	\makegapedcells
	\centering
	\caption{Classification performance of SemBC (10\% labeling), existing semi-supervised classifiers (10\% labeling), and baseline classifier (100\% labeling) indicated by mean $\pm$ standard deviation.}
	\label{tab:detailed_class_perfom}
	\resizebox{1\linewidth}{!}{
		\begin{threeparttable}
			\centering
			\begin{tabular}{l l l l l l l l l|l }		
				\toprule
				\multirow{2}{*}{Dataset}	&	\multirow{2}{*}{Metric}	&	\multicolumn{8}{c}{Model}	\\
				\cmidrule{3-10}
				&			 	&	SemBC 							&  SimCLR 				&	NNCLR				&	FixMatch			&	SunCet				& PAWS 				& Entropy			& Baseline\tnote{\dag} \\
				\midrule
				\multirow{5}{*}{MC} 		&   Accuracy  	&	$\textbf{0.946\tnote{*}}\pm0.018$		&	$0.926\pm0.033$		&	$0.926\pm0.024$		&	$0.905\pm0.032$		&	$0.927\pm0.033$		& $0.926\pm0.025$ 	& $0.917\pm0.026$	& $0.941\pm0.019$ \\
				&   Precision 	&	$\textbf{0.948\tnote{*}}\pm0.016$		&	$0.930\pm0.028$		&	$0.930\pm0.023$		&	$0.914\pm0.023$		&	$0.933\pm0.024$		& $0.931\pm0.019$	& $0.912\pm0.041$	& $0.941\pm0.016$ \\
				&   Recall 		&	$\textbf{0.947\tnote{*}}\pm0.018$		&	$0.926\pm0.034$		&	$0.927\pm0.023$		&	$0.906\pm0.032$		&	$0.927\pm0.033$		& $0.918\pm0.044$	& $0.899\pm0.059$	& $0.939\pm0.019$ \\
				&   F1-Score 	&	$\textbf{0.947\tnote{*}}\pm0.018$		&	$0.925\pm0.034$		&	$0.926\pm0.024$		&	$0.897\pm0.049$		&	$0.927\pm0.033$		& $0.915\pm0.048$	& $0.905\pm0.047$	& $0.939\pm0.018$ \\
				\midrule
				\multirow{5}{*}{BT} 		&   Accuracy  	&	$\textbf{0.877}\pm0.026$		&	$0.782\pm0.044$		&	$0.785\pm0.040$		&	$0.818\pm0.044$		&	$0.803\pm0.026$		& $0.787\pm0.034$ 	& $0.739\pm0.021$	& $\textbf{0.929\tnote{*}}\pm0.031$ \\
				&   Precision 	&	$\textbf{0.879}\pm0.028$		&	$0.789\pm0.044$		&	$0.781\pm0.041$		&	$0.847\pm0.033$		&	$0.810\pm0.024$		& $0.788\pm0.027$	& $0.774\pm0.038$	& $\textbf{0.935\tnote{*}}\pm0.022$ \\
				&   Recall 		&	$\textbf{0.879}\pm0.029$		&	$0.781\pm0.045$		&	$0.792\pm0.041$		&	$0.816\pm0.044$		&	$0.796\pm0.025$		& $0.790\pm0.035$	& $0.722\pm0.015$	& $\textbf{0.928\tnote{*}}\pm0.032$ \\
				&   F1-Score 	&	$\textbf{0.878}\pm0.027$		&	$0.774\pm0.043$		&	$0.781\pm0.052$		&	$0.813\pm0.047$		&	$0.803\pm0.027$		& $0.786\pm0.036$	& $0.735\pm0.020$	& $\textbf{0.928\tnote{*}}\pm0.032$ \\
				\midrule
				\multirow{5}{*}{SC} 		&   Accuracy  	&	$\textbf{0.929\tnote{*}}\pm0.023$		&	$0.792\pm0.080$		&	$0.744\pm0.070$		&	$0.697\pm0.059$		&	$0.775\pm0.101$		& $0.694\pm0.094$ 	& $0.654\pm0.040$	& $0.930\pm0.052$ \\
				&   Precision 	&	$\textbf{0.928}\pm0.022$		&	$0.800\pm0.071$		&	$0.744\pm0.078$		&	$0.639\pm0.109$		&	$0.756\pm0.250$		& $0.706\pm0.120$	& $0.656\pm0.153$	& $\textbf{0.938\tnote{*}}\pm0.035$ \\
				&   Recall 		&	$\textbf{0.928}\pm0.027$		&	$0.799\pm0.093$		&	$0.741\pm0.065$		&	$0.689\pm0.098$		&	$0.733\pm0.149$		& $0.679\pm0.134$	& $0.655\pm0.080$	& $\textbf{0.924\tnote{*}}\pm0.063$ \\
				&   F1-Score 	&	$\textbf{0.927\tnote{*}}\pm0.030$		&	$0.794\pm0.079$		&	$0.753\pm0.051$		&	$0.614\pm0.085$		&	$0.796\pm0.220$		& $0.699\pm0.203$	& $0.661\pm0.127$	& $0.927\pm0.061$ \\
				\midrule
				\multirow{5}{*}{CS} 		&   Accuracy  	&	$\textbf{0.987\tnote{*}}\pm0.025$		&	$0.983\pm0.037$		&	$0.983\pm0.024$		&	$0.965\pm0.033$		&	$0.963\pm0.031$		& $0.961\pm0.020$ 	& $0.950\pm0.069$	& $0.985\pm0.019$ \\
				&   Precision 	&	$\textbf{0.990}\pm0.015$		&	$0.983\pm0.035$		&	$0.979\pm0.020$		&	$0.968\pm0.028$		&	$0.976\pm0.028$		& $0.964\pm0.017$	& $0.961\pm0.053$	& $\textbf{0.990\tnote{*}}\pm0.013$ \\
				&   Recall 		&	$\textbf{0.989\tnote{*}}\pm0.017$		&	$0.983\pm0.037$		&	$0.968\pm0.024$		&	$0.965\pm0.033$		&	$0.963\pm0.031$		& $0.961\pm0.020$	& $0.948\pm0.069$	& $0.982\pm0.025$ \\
				&   F1-Score 	&	$\textbf{0.990\tnote{*}}\pm0.015$		&	$0.983\pm0.033$		&	$0.980\pm0.029$		&	$0.965\pm0.033$		&	$0.963\pm0.031$		& $0.961\pm0.020$	& $0.953\pm0.053$	& $0.982\pm0.026$ \\	
				\midrule
			\end{tabular}
			\begin{tablenotes}
				\item[\dag] A fully supervised CNN classifier.
				\item[*] The best performance.
			\end{tablenotes}
		\end{threeparttable}
	}
\end{table}
As a default setting for semi-supervised learning, we employ only 10\% of the number of labels to train our models. The classification performance, including accuracy, precision, recall, and F1-Score, of each model is shown in Table \ref{tab:detailed_class_perfom}. The proposed model (SemBC) outperforms the existing semi-supervised classifiers, such as SimCLR, NNCLR, FixMatch, SunCet, PAWS, and Entropy models. This phenomenon may occur because the existing models apply stochastic augmentation to each sample for generalization or consistency regularization purposes. Nevertheless, the important features, which are small and subtle, can disappear, as illustrated in Fig. \ref{fig:data_examples}. Accordingly, the positive and negative samples can be mixed with each other, and thus, the classifier cannot effectively discriminate them. Moreover, SemBC has a competitive performance compared with the baseline model, which is trained with a fully supervised setting (100\% labeling). The baseline model is a convolutional neural network (CNN), which combines $F_{\phi}$ and $G_{\varphi}$, as a single model (shown in Table \ref{tab:net_architecture}) for directly solving the BC problem. In the BT dataset, the baseline model significantly outperformed SemBC because the variation of the tumor and brain in MRI images was very high. Consequently, learning with a small portion of the labels makes it difficult to generalize the difference between the positive and negative samples in unseen (unlabeled) samples. Meanwhile, SimCLR, NNCLR, FixMatch, SunCet, PAWS, and Entropy had almost similar classification performances in all datasets. However, in the BT and SC datasets, their performances decreased significantly compared with our method and the baseline because the important features that distinguish the positive and negative classes are relatively smaller and more subtle compared with the MC and CS datasets. Hence, those features are relatively prone to omission by the stochastic augmentation process.

\begin{table}[h]
	\makegapedcells
	\centering
	\caption{Classification performance comparison between SemBC (10\% labeling) and baseline CNN (100\% labeling) using a transfer learning model (VGG-16), which was trained on ImageNet datasets (indicated by mean $\pm$ standard deviation).}
	\label{tab:classification_pretrained}
	\resizebox{1\linewidth}{!}{
		\centering
		\begin{tabular}{l l l l l|l l l l} 		
			\toprule
			\multirow{2}{*}{Model}		&	\multicolumn{4}{c|}{MC}																		&	\multicolumn{4}{c}{BT}																		\\
			\cmidrule{2-9}
			&	Accuracy			&	Precision			&	Recall 				&	F1-Score			&	Accuracy			&	Precision			&	Recall 				&	F1-Score			\\
			\midrule
			Baseline  						&	$0.945\pm0.004$		&	$0.947\pm0.004$		&	$0.945\pm0.004$		&	$0.945\pm0.004$		&	$0.984\pm0.006$		&	$0.984\pm0.007$		&	$0.984\pm0.006$		&	$0.984\pm0.008$		\\
			SemBC  					&	$0.946\pm0.004$		&	$0.946\pm0.004$		&	$0.946\pm0.004$		&	$0.946\pm0.003$		&	$0.927\pm0.010$		&	$0.927\pm0.010$		&	$0.926\pm0.009$		&	$0.927\pm0.010$		\\
			\midrule
			\multirow{2}{*}{Model}		&	\multicolumn{4}{c|}{SC}																		&	\multicolumn{4}{c}{CS}																		\\
			\cmidrule{2-9}
			&	Accuracy			&	Precision			&	Recall 				&	F1-Score			&	Accuracy			&	Precision			&	Recall 				&	F1-Score			\\
			\midrule
			Baseline						&	$0.996\pm0.002$ 	&	$0.995\pm0.002$		&	$0.994\pm0.001$		&	$0.996\pm0.001$		&	$0.995\pm0.005$ 	&	$0.996\pm0.004$		&	$0.995\pm0.005$		&	$0.995\pm0.005$		\\
			SemBC					&	$0.995\pm0.001$ 	&	$0.994\pm0.003$		&	$0.996\pm0.001$		&	$0.995\pm0.002$		&	$0.994\pm0.005$ 	&	$0.995\pm0.004$		&	$0.994\pm0.004$		&	$0.995\pm0.004$		\\
			\bottomrule
		\end{tabular}
	}
\end{table}

In addition, we implemented a transfer learning mechanism to assess SemBC performance against the baseline. In the present study, we employed the VGG-16 model, which was trained with the ImageNet dataset, as the backbone network in our encoder network ($F_{\phi}$). Meanwhile, in the baseline model, we directly utilized the VGG-16 on the top of the classifier, similar to previous works \cite{rajaraman2018pre}, \cite{s22010372}, \cite{DEITSCH2019455}, \cite{s22103662}. As shown in Table \ref{tab:classification_pretrained}, the baseline model yields an accuracy result similar to that reported in \cite{rajaraman2018pre} (MC), \cite{s22010372} (BT), \cite{DEITSCH2019455} (SC), \cite{s22103662} (CS). It is notable that SemBC has a higher classification performance in the BT dataset than the SemBC with the basic training (shown in Table \ref{tab:detailed_class_perfom}). Nonetheless, this result is still not significantly competitive compared to the baseline accuracy. Therefore, given the limited supervision (10\% labeling), fine-tuning with transfer learning still suffers from a generalization error in encountering high variation of binary samples, as in the BT dataset.

\subsection{Model properties}
\begin{figure}
	\centering
	\begin{tikzpicture}[scale=0.30,font=\large]
		\begin{axis}[
			title=\LARGE{MC},
			width  = 0.8*\textwidth,
			height = 7.5cm,
			xbar=2*\pgflinewidth,
			bar width=8pt,
			symbolic y coords={Accuracy,Precision,Recall,F1-Score},
			ytick = data,scaled x ticks = false, 
			enlarge y limits=0.1,
			nodes near coords={\pgfmathprintnumber\pgfplotspointmeta},
			nodes near coords align={horizontal},
			every node near coord/.append style={font=\small,xshift=18pt,/pgf/number format/.cd,precision=3},
			xmin = 0.87,
			xmax = 0.98,
			legend image code/.code={%
				\draw[#1, draw=none] (0cm,-0.1cm) rectangle (0.65cm,0.22cm);
			},  
			legend style={at={(0.5,-0.17)},
				anchor=north,legend columns=-1},
			every axis plot/.append style={fill opacity=1.0},
			every tick label/.append style={font=\Large}
			]
			\addplot[xbar, black,fill=red!50,postaction={pattern=north east lines},error bars/.cd,
			x dir=both,x explicit] coordinates {
				(0.946,{Accuracy}) += (0.005770193,0) -=(0.005770193,0)
				(0.948,{Precision}) += (0.005118808,0) -=(0.005118808,0)
				(0.947,{Recall}) += (0.005755989,0) -=(0.005755989,0)
				(0.947,{F1-Score}) += (0.005794601,0) -=(0.005794601,0) 
			};
			\addlegendentry{$m$=10\%}
			\addplot[xbar,black,fill=black,postaction={pattern=horizontal lines},error bars/.cd,
			x dir=both,x explicit] coordinates {
				(0.946,{Accuracy}) += (0.005453966,0) -=(0.005453966,0)
				(0.949,{Precision}) += (0.005435035,0) -=(0.005435035,0)
				(0.947,{Recall}) += (0.005123533,0) -=(0.005123533,0)
				(0.947,{F1-Score}) += (0.005794601,0) -=(0.005794601,0)
			};
			\addlegendentry{$m$=30\%}
			\addplot[xbar,black,fill=green!50,postaction={pattern=dots},error bars/.cd,
			x dir=both,x explicit] coordinates {
				(0.946,{Accuracy}) += (0.00482151,0) -=(0.00482151,0)
				(0.949,{Precision}) += (0.005118808,0) -=(0.005118808,0)
				(0.948,{Recall}) += (0.005439761,0) -=(0.005439761,0)
				(0.947,{F1-Score}) += (0.005162146,0) -=(0.005162146,0)
				
			};
			\addlegendentry{$m$=50\%}
		\end{axis}
	\end{tikzpicture}
	\begin{tikzpicture}[scale=0.30,font=\large]
		\begin{axis}[
			title=\LARGE{BT},
			width  = 0.8*\textwidth,
			height = 7.5cm,
			xbar=2*\pgflinewidth,
			bar width=8pt,
			symbolic y coords={Accuracy,Precision,Recall,F1-Score},
			ytick = data,scaled x ticks = false, 
			enlarge y limits=0.1,
			nodes near coords={\pgfmathprintnumber\pgfplotspointmeta},
			nodes near coords align={horizontal},
			every node near coord/.append style={font=\small,xshift=14pt,/pgf/number format/.cd,precision=3},
			xmax = 0.99,
			legend image code/.code={%
				\draw[#1, draw=none] (0cm,-0.1cm) rectangle (0.65cm,0.22cm);
			},  
			legend style={at={(0.5,-0.17)},
				anchor=north,legend columns=-1},
			every axis plot/.append style={fill opacity=1.0},
			every tick label/.append style={font=\Large}
			]
			\addplot[xbar, black,fill=red!50,postaction={pattern=north east lines},error bars/.cd,
			x dir=both,x explicit] coordinates {
				(0.879,{Accuracy}) += (0.008913284,0) -=(0.008913284,0)
				(0.879,{Precision}) += (0.00896714,0) -=(0.00896714,0) 
				(0.879,{Recall}) += (0.008811249,0) -=(0.008811249,0)
				(0.878,{F1-Score}) += (0.008925838,0) -=(0.008925838,0)
			};
			\addlegendentry{$m$=10\%}
			\addplot[xbar,black,fill=black,postaction={pattern=horizontal lines},error bars/.cd,
			x dir=both,x explicit] coordinates {
				(0.949,{Accuracy}) += (0.003497795,0) -=(0.003497795,0)
				(0.950,{Precision}) += (0.003580099,0) -=(0.003580099,0)
				(0.949,{Recall}) += (0.003431007,0) -=(0.003431007,0)
				(0.949,{F1-Score}) += (0.003495558,0) -=(0.003495558,0)
			};
			\addlegendentry{$m$=30\%}
			\addplot[xbar,black,fill=green!50,postaction={pattern=dots},error bars/.cd,
			x dir=both,x explicit] coordinates {
				(0.964,{Accuracy}) += (0.00259451,0) -=(0.00259451,0)
				(0.964,{Precision}) += (0.002512422,0) -=(0.002512422,0)
				(0.964,{Recall}) += (0.002518481,0) -=(0.002518481,0)
				(0.964,{F1-Score}) += (0.002604617,0) -=(0.002604617,0)
				
			};
			\addlegendentry{$m$=50\%}
		\end{axis}
	\end{tikzpicture}
	\begin{tikzpicture}[scale=0.30,font=\large]
		\begin{axis}[
			title=\LARGE{SC},
			width  = 0.8*\textwidth,
			height = 7.5cm,
			xbar=2*\pgflinewidth,
			bar width=8pt,
			symbolic y coords={Accuracy,Precision,Recall,F1-Score},
			ytick = data,scaled x ticks = false, 
			enlarge y limits=0.1,
			nodes near coords={\pgfmathprintnumber\pgfplotspointmeta},
			nodes near coords align={horizontal},
			every node near coord/.append style={font=\small,xshift=18pt,/pgf/number format/.cd,precision=3},
			xmin = 0.9,
			xmax = 1.0,
			legend image code/.code={%
				\draw[#1, draw=none] (0cm,-0.1cm) rectangle (0.65cm,0.22cm);
			},  
			legend style={at={(0.5,-0.17)},
				anchor=north,legend columns=-1},
			every axis plot/.append style={fill opacity=1.0},
			every tick label/.append style={font=\Large}
			]
			\addplot[xbar, black,fill=red!50,postaction={pattern=north east lines},error bars/.cd,
			x dir=both,x explicit] coordinates {
				(0.929,{Accuracy}) += (0.007350872,0) -=(0.007350872,0)
				(0.928,{Precision}) += (0.00689369,0) -=(0.00689369,0)
				(0.928,{Recall}) += (0.00847671,0) -=(0.00847671,0)
				(0.927,{F1-Score}) += (0.009411458,0) -=(0.009411458,0)
			};
			\addlegendentry{$m$=10\%}
			\addplot[xbar,black,fill=black,postaction={pattern=horizontal lines},error bars/.cd,
			x dir=both,x explicit] coordinates {
				(0.969,{Accuracy}) += (0.00730445,0) -=(0.00730445,0)
				(0.969,{Precision}) += (0.007589928,0) -=(0.007589928,0)
				(0.968,{Recall}) += (0.007428423,0) -=(0.007428423,0)
				(0.969,{F1-Score}) += (0.007507464,0) -=(0.007507464,0)
			};
			\addlegendentry{$m$=30\%}
			\addplot[xbar,black,fill=green!50,postaction={pattern=dots},error bars/.cd,
			x dir=both,x explicit] coordinates {
				(0.972,{Accuracy}) += (0.009811709,0) -=(0.009811709,0) 
				(0.974,{Precision}) += (0.010522166,0) -=(0.010522166,0)
				(0.975,{Recall}) += (0.010406467,0) -=(0.010406467,0)
				(0.979,{F1-Score}) += (0.010399946,0) -=(0.010399946,0)
				
			};
			\addlegendentry{$m$=50\%}
		\end{axis}
	\end{tikzpicture}
	\begin{tikzpicture}[scale=0.30,font=\large]
		\begin{axis}[
			title=\LARGE{CS},
			width  = 0.8*\textwidth,
			height = 7.5cm,
			xbar=2*\pgflinewidth,
			bar width=8pt,
			symbolic y coords={Accuracy,Precision,Recall,F1-Score},
			ytick = data,scaled x ticks = false, 
			enlarge y limits=0.1,
			nodes near coords={\pgfmathprintnumber\pgfplotspointmeta},
			nodes near coords align={horizontal},
			every node near coord/.append style={font=\small,xshift=18pt,/pgf/number format/.cd,precision=3},
			xmin = 0.94,
			xmax = 1.01,
			legend image code/.code={%
				\draw[#1, draw=none] (0cm,-0.1cm) rectangle (0.65cm,0.22cm);
			},  
			legend style={at={(0.5,-0.17)},
				anchor=north,legend columns=-1},
			every axis plot/.append style={fill opacity=1.0},
			every tick label/.append style={font=\Large}
			]
			\addplot[xbar, black,fill=red!50,postaction={pattern=north east lines},error bars/.cd,
			x dir=both,x explicit] coordinates {
				(0.988,{Accuracy}) += (0.00793531,0) -=(0.00793531,0)
				(0.990,{Precision}) += (0.004699647,0) -=(0.004699647,0)
				(0.989,{Recall}) += (0.005284035,0) -=(0.005284035,0)
				(0.990,{F1-Score}) += (0.004773633,0) -=(0.004773633,0)
			};
			\addlegendentry{$m$=10\%}
			\addplot[xbar,black,fill=black,postaction={pattern=horizontal lines},error bars/.cd,
			x dir=both,x explicit] coordinates {
				(0.989,{Accuracy}) += (0.007941141,0) -=(0.007941141,0)
				(0.989,{Precision}) += (0.004103011,0) -=(0.004103011,0)
				(0.990,{Recall}) += (0.003903424,0) -=(0.003903424,0) 
				(0.990,{F1-Score}) += (0.003825298,0) -=(0.003825298,0)
			};
			\addlegendentry{$m$=30\%}
			\addplot[xbar,black,fill=green!50,postaction={pattern=dots},error bars/.cd,
			x dir=both,x explicit] coordinates {
				(0.993,{Accuracy}) += (0.00537382,0) -=(0.00537382,0)
				(0.990,{Precision}) += (0.005625466,0) -=(0.005625466,0)
				(0.994,{Recall}) += (0.005481293,0) -=(0.005481293,0)
				(0.990,{F1-Score}) += (0.004447597,0) -=(0.004447597,0)
				
			};
			\addlegendentry{$m$=50\%}
		\end{axis}
	\end{tikzpicture}
	\caption{Effect of increasing the number of labels ($m\%$) on model classification performance.}\label{fig:comparison_acc_knn}
\end{figure}
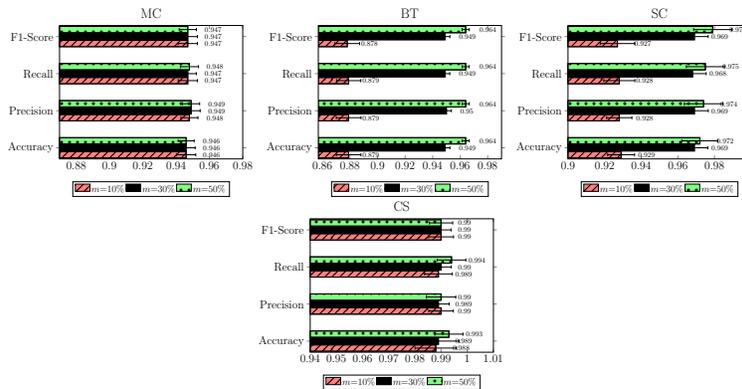

In addition to using 10\% labeling, in this section, we increase the proportion of labeling to determine its impact on classification performance. We denote $m$ as a portion of the labeled samples used during the training process. Ideally, by increasing the $m$ value, the model can learn more, such that the classification performance can be improved, as shown in the BT and SC datasets in Table \ref{fig:comparison_acc_knn}. The classification performance significantly increased from $m=10\%$ to $m=30\%$ and increased less when $m>30\%$. For the BT dataset, we can conclude that adding more labels is more effective than fine-tuning with transfer learning, as shown in Table \ref{tab:classification_pretrained}. Thus, it is an evident that the data variability of BT, which distinguishes between positive and negative classes in brain MRI, is sufficiently high. Conversely, in MC and CS, increasing the value of $m$ does not significantly change the accuracy performance. This can occur because the model has already attained the optimum decision boundary in discriminating the positive and negative samples; thus, the model performance becomes saturated even by adding more data. It is notable that the classification performances of MC and CS have approximately similar results to those of the baseline, as shown in Tables \ref{tab:detailed_class_perfom} and \ref{tab:classification_pretrained}. Therefore, if the dataset is separable enough using semi-supervised learning with a small portion of labeled data, it is sufficient to attain nearly optimal results. However, if the data are sufficiently complex, such as those with high inseparability or variability, more labels are required to achieve nearly optimal results.

\begin{figure}[h]
	\centering
	\subfigure[MC]{\includegraphics[width=0.28\linewidth]{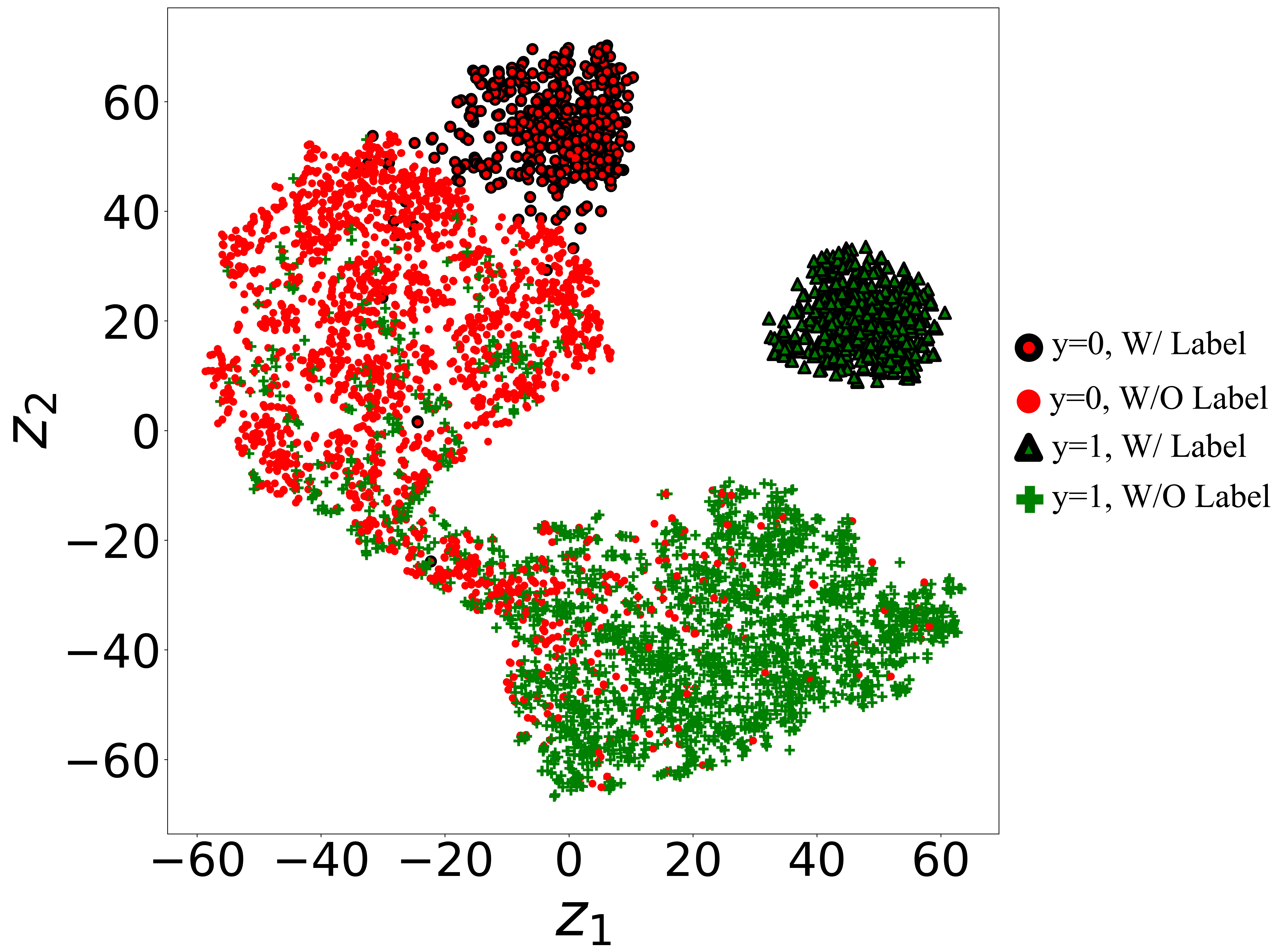}
		\label{cell_latent_vis}}
	\subfigure[BT]{\includegraphics[width=0.28\linewidth]{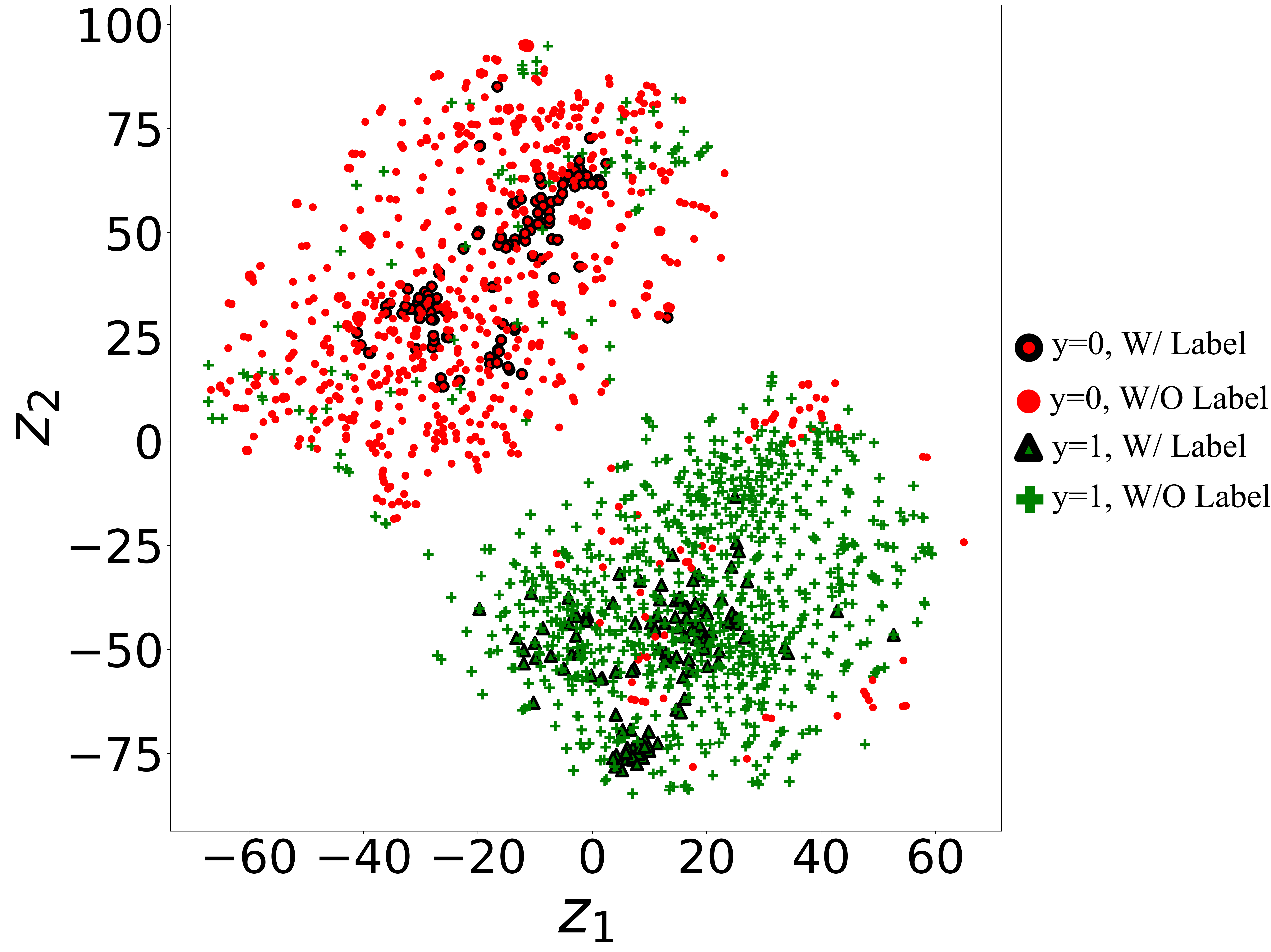}
		\label{tumor_latent_vis}}
	\subfigure[SC]{\includegraphics[width=0.28\linewidth]{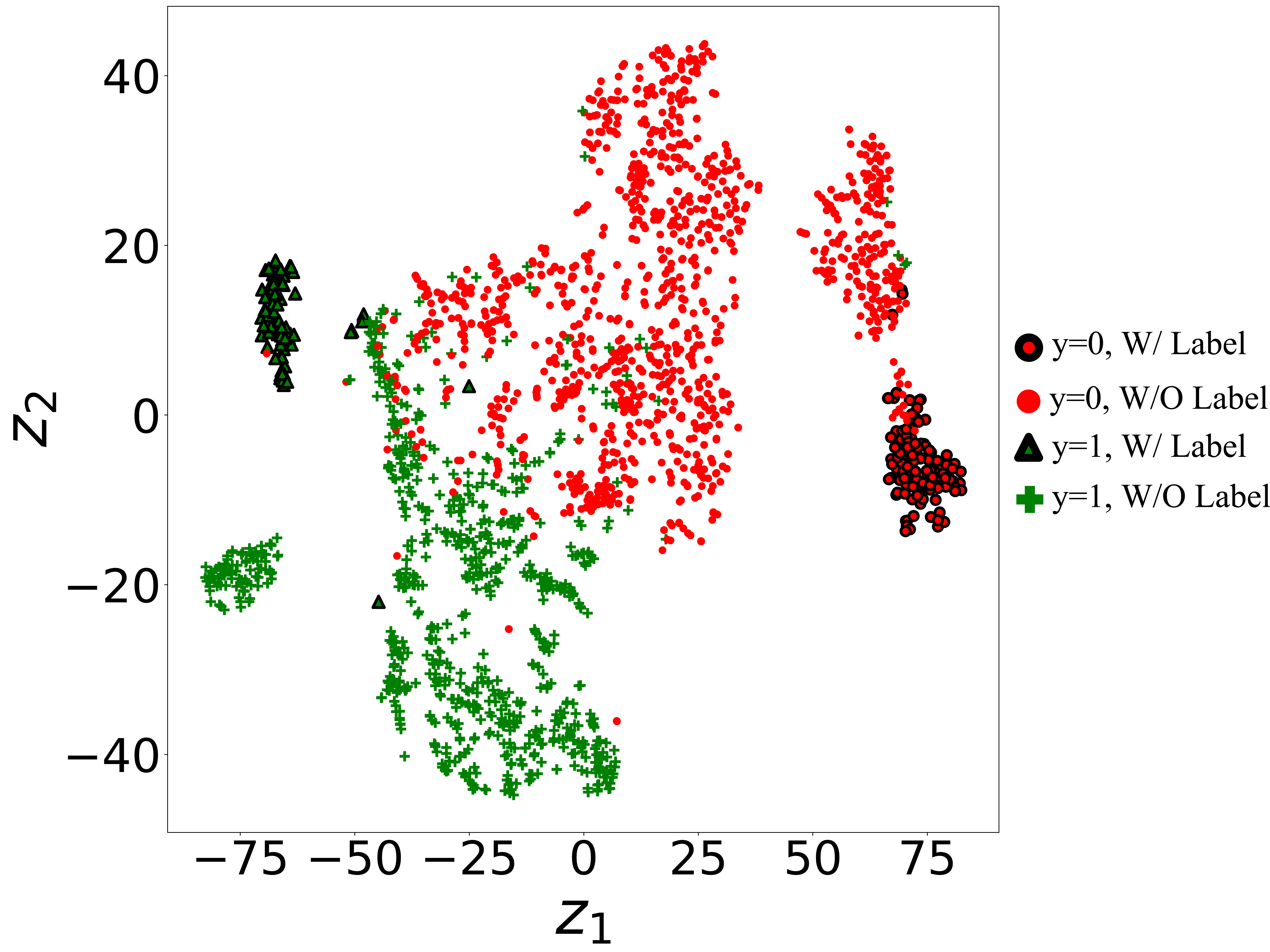}
		\label{solar_latent_vis}}
	\subfigure[CS]{\includegraphics[width=0.28\linewidth]{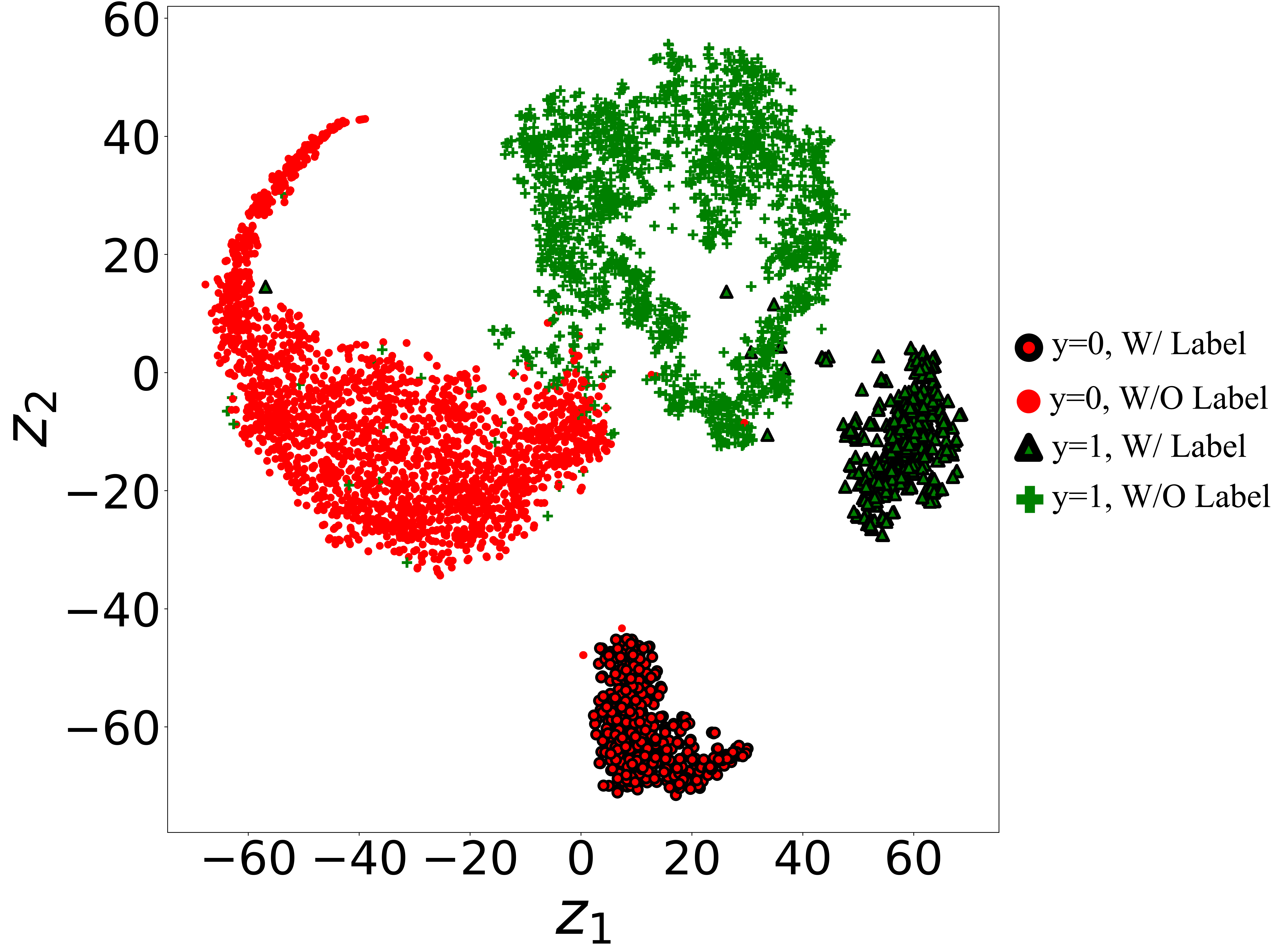}
		\label{crack_latent_vis}}
	\caption{Visualization of binary manifold representation learned using an angular distance of labeled (10\%) samples.}
	\label{fig:latent_vis}
\end{figure}

The manifold representation generated by $F_{\phi}$ with labeled samples is generally devised to obtain positive and negative class positions based on the angular distance. Ideally, positive and negative samples should be far apart from each other. Hence, we can easily determine the location of any unlabeled sample in the manifold representation based on its proximity to either positive or negative samples. To observe this situation, we projected the manifold representation onto a two-dimensional ($2D$) space using the t-SNE algorithm \cite{KAMAL2022108562}. It is noteworthy that the $F_{\phi}$ results in a vector with length 256, as shown in Table \ref{tab:net_architecture}; thus, converting it into a $2D$ space may yield an unavoided estimation error. We utilized only 10\% of the labels to define manifold spaces and place the location of the unlabeled samples. The estimated $2D$ visualization is shown in Fig. \ref{fig:latent_vis}. We can infer that the CS dataset is more separable than the other datasets. Thus, its classification performance is high, as shown in Table \ref{tab:detailed_class_perfom}. In the BT dataset, the labeled samples tend to be more scattered (less dense) than those in the other datasets. This can occur because the brain MRI samples have high variability. Meanwhile, because some samples in the MC and SC have a subtle difference between positive and negative samples, devising an ideal decision boundary to separate positive and negative classes in the SC and MC datasets becomes more challenging than in other datasets.

\begin{figure}
	\centering
	\begin{tikzpicture}[scale=0.29,font=\large]
		\begin{axis}[
			ybar,
			title=\LARGE{MC},
			legend style={at={(0.01,0.98)},anchor=north west},
			symbolic x coords={$k=3$,$k=11$,$k=19$},
			bar width=8pt,bar shift=-0.2cm,
			legend entries={accuracy},
			legend image code/.code={%
				\draw[#1] (0cm,-0.1cm) rectangle (0.3cm,0.1cm);
			}  
			height = 7.5cm,
			xtick=data,
			axis y line*=left,
			ylabel=accuracy,
			ymin = 0.87,
			ymax = 0.98,
			every tick label/.append style={font=\Large}
			]
			\addplot[black,fill=red!50,postaction={pattern=north east lines},error bars/.cd,
			y dir=both,
			y explicit relative] coordinates {
				($k=3$,0.938268917) +- (0,0.005543908) 
				($k=11$,0.945724732) +- (0,0.005770193)
				($k=19$,0.948277082) +- (0,0.005475258)
			};
		\end{axis}
		\begin{axis}[
			ybar,
			legend style={at={(0.01,0.88)},anchor=north west},
			symbolic x coords={$k=3$,$k=11$,$k=19$},
			bar width=8pt,bar shift=0.2cm,
			legend entries={time},
			legend image code/.code={%
				\draw[#1] (0cm,-0.1cm) rectangle (0.3cm,0.1cm);
			}
			height = 7.5cm,
			xtick=data,
			ylabel=seconds,
			axis y line*=right,
			ymin = 1,
			ymax = 220,
			every tick label/.append style={font=\Large}
			]
			\addplot[black,fill=green!50,error bars/.cd,
			y dir=both,
			y explicit relative] coordinates {
				($k=3$, 42.74864157) +- (0,0.0179213872)
				($k=11$, 120.8841467) +- (0,0.0193403171)
				($k=19$, 206.2625709) +- (0,0.01800464576)
			};
		\end{axis}
	\end{tikzpicture}
	\begin{tikzpicture}[scale=0.29,font=\large]
		\begin{axis}[
			ybar,
			title=\LARGE{BT},
			legend style={at={(0.01,0.98)},anchor=north west},
			symbolic x coords={$k=3$,$k=11$,$k=19$},
			bar width=8pt,bar shift=-0.2cm,
			legend entries={accuracy},
			legend image code/.code={%
				\draw[#1] (0cm,-0.1cm) rectangle (0.3cm,0.1cm);
			}  
			height = 7.5cm,
			xtick=data,
			axis y line*=left,
			ylabel=accuracy,
			ymin = 0.80,
			ymax = 0.935,
			every tick label/.append style={font=\Large}
			]
			\addplot[black,fill=red!50,postaction={pattern=north east lines},error bars/.cd,
			y dir=both,
			y explicit relative] coordinates {
				($k=3$,0.877333333) +- (0,0.008991222) 
				($k=11$,0.878666667) +- (0,0.008913284)
				($k=19$,0.8875) +- (0,0.006670453)
			};
		\end{axis}
		\begin{axis}[
			ybar,
			legend style={at={(0.01,0.88)},anchor=north west},
			symbolic x coords={$k=3$,$k=11$,$k=19$},
			bar width=8pt,bar shift=0.2cm,
			legend entries={time},
			legend image code/.code={%
				\draw[#1] (0cm,-0.1cm) rectangle (0.3cm,0.1cm);
			}
			height = 7.5cm,
			xtick=data,
			ylabel=seconds,
			axis y line*=right,
			ymin = 1,
			ymax = 200,
			every tick label/.append style={font=\Large}
			]
			\addplot[black,fill=green!50,error bars/.cd,
			y dir=both,
			y explicit relative] coordinates {
				($k=3$, 52.45068844) +- (0,0.01745382)
				($k=11$, 115.9329515) +- (0,0.017243401)
				($k=19$, 182.7382586) +- (0,0.015667576)
			};
		\end{axis}
	\end{tikzpicture}
	\begin{tikzpicture}[scale=0.29,font=\large]
		\begin{axis}[
			ybar,
			title=\LARGE{SC},
			legend style={at={(0.01,0.98)},anchor=north west},
			symbolic x coords={$k=3$,$k=11$,$k=19$},
			bar width=8pt,bar shift=-0.2cm,
			legend entries={accuracy},
			legend image code/.code={%
				\draw[#1] (0cm,-0.1cm) rectangle (0.3cm,0.1cm);
			}  
			height = 7.5cm,
			xtick=data,
			axis y line*=left,
			ylabel=accuracy,
			ymin = 0.86,
			ymax = 0.96,
			every tick label/.append style={font=\Large}
			]
			\addplot[black,fill=red!50,postaction={pattern=north east lines},error bars/.cd,
			y dir=both,
			y explicit relative] coordinates {
				($k=3$,0.921755725) +- (0,0.008838577) 
				($k=11$,0.929007634) +- (0,0.007350872)
				($k=19$,0.928937405) +- (0,0.006699689)
			};
		\end{axis}
		\begin{axis}[
			ybar,
			legend style={at={(0.01,0.88)},anchor=north west},
			symbolic x coords={$k=3$,$k=11$,$k=19$},
			bar width=8pt,bar shift=0.2cm,
			legend entries={time},
			legend image code/.code={%
				\draw[#1] (0cm,-0.1cm) rectangle (0.3cm,0.1cm);
			}
			height = 7.5cm,
			xtick=data,
			ylabel=seconds,
			axis y line*=right,
			ymin = 0.0,
			ymax = 20,
			every tick label/.append style={font=\Large}
			]
			\addplot[black,fill=green!50,error bars/.cd,
			y dir=both,
			y explicit relative] coordinates {
				($k=3$, 4.984485149) +- (0,0.014340792)
				($k=11$, 11.19044092) +- (0,0.01429871)
				($k=19$, 17.76781331) +- (0,0.014347886)
			};
		\end{axis}
	\end{tikzpicture}
	\begin{tikzpicture}[scale=0.29,font=\large]
		\begin{axis}[
			ybar,
			title=\LARGE{CS},
			legend style={at={(0.01,0.98)},anchor=north west},
			symbolic x coords={$k=3$,$k=11$,$k=19$},
			bar width=8pt,bar shift=-0.2cm,
			legend entries={accuracy},
			legend image code/.code={%
				\draw[#1] (0cm,-0.1cm) rectangle (0.3cm,0.1cm);
			}  
			height = 7.5cm,
			xtick=data,
			axis y line*=left,
			ylabel=accuracy,
			ymin = 0.94,
			ymax = 1.01,
			every tick label/.append style={font=\Large}
			]
			\addplot[black,fill=red!50,postaction={pattern=north east lines},error bars/.cd,
			y dir=both,
			y explicit relative] coordinates {
				($k=3$,0.98) +- (0,0.00424871) 
				($k=11$,0.99) +- (0,0.004773033)
				($k=19$,0.989) +- (0,0.005196659)
			};
		\end{axis}
		\begin{axis}[
			ybar,
			legend style={at={(0.01,0.88)},anchor=north west},
			symbolic x coords={$k=3$,$k=11$,$k=19$},
			bar width=8pt,bar shift=0.2cm,
			legend entries={time},
			legend image code/.code={%
				\draw[#1] (0cm,-0.1cm) rectangle (0.3cm,0.1cm);
			}
			height = 7.5cm,
			xtick=data,
			ylabel=seconds,
			axis y line*=right,
			ymin = 10,
			ymax = 320,
			every tick label/.append style={font=\Large}
			]
			\addplot[black,fill=green!50,error bars/.cd,
			y dir=both,
			y explicit relative] coordinates {
				($k=3$, 62.3829773) +- (0,0.0179213872)
				($k=11$, 180.1571909) +- (0,0.0193403171)
				($k=19$, 310.6456102) +- (0,0.01800464576)
			};
		\end{axis}
	\end{tikzpicture}
	\caption{Effect of varying numbers of $k$ in defining on-the-fly labels on the accuracy performance and learning time (seconds).}\label{fig:accuracy_vs_time}
\end{figure}
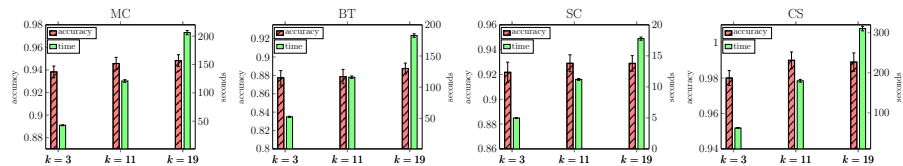

In addition, we observed the impact of varying $k$ on classifier accuracy and learning time. The $k$-random samples were used to define the on-the-fly label. As shown in Fig. \ref{fig:accuracy_vs_time}, by increasing $k$, the classifier accuracy tends to improve less significantly and even saturates after $k=11$. In contrast, the learning time linearly increases with an increasing value of $k$. Hence, as a nearest neighbor-based method, the proposed model suffers from high computational cost with an increasing $k$ value; however, this impact is not very significant to the classifier accuracy. Thus, using a small $k$ value is sufficient to obtain satisfactory results.

\begin{table}
	\makegapedcells
	\centering
	\caption{Training time (seconds) comparison for each models in BT dataset. Mean $\pm$ standard deviation.}
	\label{tab:detailed_classfication_acc}
	\resizebox{0.87\linewidth}{!}{
		\begin{threeparttable}
			\centering
			\begin{tabular}{l l l l l l l l}		
				\toprule
				\multirow{2}{*}{Component}	&	\multicolumn{7}{c}{Model}	\\
				\cmidrule{2-8}
				&	SemBC 					&  SimCLR 				&	NNCLR				&	FixMatch			&	SunCet				& PAWS 				& Entropy		 \\
				\midrule
				$F_{\phi}$ 					&	$20.796\pm0.166$		&	$2.801\pm0.056$		&	$11.949\pm0.161$	&	$1.512\pm0.006$		&	$2.626\pm0.028$		& $2.865\pm0.004$	& $1.503\pm0.014$	\\
				$G_{\varphi}$ 				&	$50.032\pm0.508$		&	$1.423\pm0.017$		&	$1.471\pm0.027$		&	$23.807\pm0.241$	&	$1.496\pm0.020$		& $1.430\pm0.007$	& $1.485\pm0.021$	\\
				\midrule
				Total						&	$70.828$				&	$4.224$				&	$13.420$				&	$25.319$			&	$4.123$				& $4.295$			& $2.988$	\\
				\midrule
			\end{tabular}
		\end{threeparttable}
	}
\end{table}

Finally, we conducted a training time (per epoch) comparison for each semi-supervised model. There are two primary components for each model, i.e., the encoder ($F_{\phi}$) and binary classifier network ($G_{\phi}$). As shown in Table \ref{tab:detailed_classfication_acc}, the proposed method has the longest training time for both $F_{\phi}$ and $G_{\phi}$ components. In $F_{\phi}$, SemBC, which is based on the Siamese setting \cite{Chicco2021}, has a different learning representation than other models. The input data are doubled, and the pairwise distance between matching and non-matching is calculated; thus, there are two feedforward processes in a single iteration (shown in Algorithms \ref{algo:binary_distance} and \ref{algo:mapping_pair}). In NNCLR, besides the nearest neighbor-based representation, a correlation matrix is used to compute the similarity; therefore, it has a high computational cost in $F_{\phi}$. FixMatch and Entropy have a low computational cost in $F_{\phi}$ because they merely use labeled samples in this component. In contrast, the computation time of FixMatch is high in the $G_{\varphi}$ because of the pseudo-labeling mechanism. Our SemBC has the highest computation time in $G_{\varphi}$ because we use $k$ samples in both positive and negative classes to define the on-the-fly label. Thus, we conclude that although SemBC has a superior classification performance compared with state-of-the-art semi-supervised classifiers, it still has a limitation in terms of computational cost. Nevertheless, once the training is complete, all these methods require similar computation time in the testing phase.

\section{Conclusion}
\label{sec:conclusion}

Several significant semi-supervised learning methods, which rely heavily on stochastic data augmentation, have been reported to effectively solve the multi-class classification problem. However, we demonstrated that stochastic data augmentation is not suitable for a typical BC problem because it can reduce or even remove region-of-interest (ROI) features. In BC datasets, the ROI is typically small and subtle. The smaller (or more subtle) the ROI, the higher the classification error will be. Unlike previous studies, we introduced a distance-based method to solve the BC problem. The encoder is trained with few labels to define binary manifold representation using metric learning, whereas the classifier utilizes an on-the-fly label during training to approximate the labels of unlabeled samples based on manifold representation. The results indicate that the proposed method outperforms state-of-the-art semi-supervised classifiers, such as SimCLR, NNCLR, FixMatch, SunCet, PAWS, and Entropy. Moreover, with only 10\% of the labels and without any data augmentation technique, the proposed method achieves a result similar to that of a fully supervised classifier. Therefore, the distance-based technique is a promising approach for solving various BC or anomaly detection problems with limited or less supervision in medical testing, quality control, information retrieval, and so on. Nonetheless, despite its superior performance, we acknowledge that our method still suffers from a higher computational cost during training than existing methods. Accordingly, the computation time could increase if we apply it to the multi-class classification problem along with an increasing number of classes. Therefore, this is a new research direction that needs to be addressed in the future.

\section*{CRediT authorship contribution statement}
\textbf{Imam Mustafa Kamal:} Conceptualization, Methodology, Software, Validation, Formal analysis, Investigation, Writing – original draft, Writing – review \& editing, Visualization. \textbf{Hyerim Bae:} Writing – review \& editing, Supervision, Funding acquisition.

\section*{Declaration of competing interest}
The authors declare that they have no known competing financial interests or personal relationships that could have appeared to influence the work reported in this paper.

\section*{Acknowledgements}
This research was supported by the MSIT(Ministry of Science and ICT), Korea, under the Grand Information Technology Research Center support program(IITP-2022-2020-0-01791) supervised by the IITP(Institute for Information \& communications Technology Planning \& Evaluation).

\bibliographystyle{unsrtnat}
\bibliography{texfile}

\end{document}